\documentclass{bmvc2k}

\usepackage{amsfonts}
\usepackage{mathrsfs}
\usepackage{siunitx}
\usepackage{booktabs}
\usepackage{multirow}
\usepackage[table]{xcolor}
\usepackage{xspace}
\usepackage{pifont}
\usepackage{float}
\usepackage{hyperref}

\newcommand{\bmvc}[1]{{\color{black}#1}}
\newcommand{\bmv}[1]{{\color{black}#1}}
\newcommand{\ready}[1]{{\color{black}#1}}
\newcommand{\name}[0]{TQD-Track}
\newcommand{\lt}{\ensuremath <}

\newcommand{\ie}{i.e.\@\xspace}

\title{TQD-Track: Temporal Query Denoising for 3D Multi-Object Tracking}

\addauthor{Yutong Yang$^*$}{yutong.yang@mercedes-benz.com}{1,3}
\addauthor{Shuxiao Ding$^*$}{dingshuxiao0511@gmail.com}{2}
\addauthor{Mohammed Amine Bencheikh Lehocine}{mohammed.lehocine@mercedes-benz.com}{1}
\addauthor{Julian Wiederer}{julian.wiederer@mercedes-benz.com}{1}
\addauthor{Markus Braun}{braun.markus89@gmail.com}{1}
\addauthor{Peizheng Li}{https://edwardleelpz.github.io/}{1,4}
\addauthor{Juergen Gall}{https://pages.iai.uni-bonn.de/gall_juergen/}{2,5}
\addauthor{Bin Yang}{https://www.iss.uni-stuttgart.de/}{3}

\addinstitution{Mercedes-Benz AG}
\addinstitution{University of Bonn}
\addinstitution{University of Stuttgart}
\addinstitution{University of T\"ubingen}
\addinstitution{Lamarr Institute for Machine Learning and Artificial Intelligence}

\runninghead{Yang, Ding, et al.}{TQD-Track}

\makeatletter
\let\BMVA@origmaketitle\maketitle
\renewcommand{\maketitle}{%
  \let\BMVA@origurl\url
  \renewcommand{\url}[1]{%
    \begingroup
    \edef\BMVA@thisurl{\detokenize{##1}}%
    \edef\BMVA@shortemail{\detokenize{mohammed.lehocine@mercedes-benz.com}}%
    \ifx\BMVA@thisurl\BMVA@shortemail
      \endgroup
      \href{mailto:mohammed_amine.bencheikh_lehocine@mercedes-benz.com}{\nolinkurl{mohammed.lehocine@mercedes-benz.com}}%
    \else
      \endgroup
      \BMVA@origurl{##1}%
    \fi
  }%
  \BMVA@origmaketitle
  \let\url\BMVA@origurl
}
\makeatother

\def\eg{\emph{e.g}\bmvaOneDot}

\makeatletter
\let\BMVA@origblfootnote\BMVA@blfootnote
\renewcommand\BMVA@blfootnote[1]{%
  \BMVA@origblfootnote{\null\hspace{-1.9em}$^*$Equal contribution. Shuxiao Ding and Markus Braun contributed to this work while at Mercedes-Benz AG.\\\hspace*{1.9em}#1}%
}
\makeatother

\newcommand{\supplementrunninghead}{\emph{YANG, DING, ET AL.: TQD-TRACK}}
\makeatletter
\def\ps@supplement{%
  \def\@oddhead{%
    \vbox{%
      \hbox to\textwidth{\small\supplementrunninghead\hfil\small\bfseries\thepage}%
      \vskip 4pt%
      \hrule height 0.4pt%
    }%
  }%
  \def\@evenhead{%
    \vbox{%
      \hbox to\textwidth{\small\supplementrunninghead\hfil\small\bfseries\thepage}%
      \vskip 4pt%
      \hrule height 0.4pt%
    }%
  }%
  \def\@oddfoot{}%
  \def\@evenfoot{}%
}
\makeatother

\newcommand{\makesupplementheader}{%
  \clearpage%
  \setcounter{page}{1}%
  \setcounter{section}{0}%
  \setcounter{figure}{0}%
  \setcounter{table}{0}%
  \setcounter{equation}{0}%
  \renewcommand{\thepage}{\arabic{page}}%
  \renewcommand{\thesection}{A\arabic{section}}%
  \renewcommand{\thesubsection}{A\arabic{section}.\arabic{subsection}}%
  \renewcommand{\thesubsubsection}{A\arabic{section}.\arabic{subsection}.\arabic{subsubsection}}%
  \renewcommand{\thefigure}{A\arabic{figure}}%
  \renewcommand{\thetable}{A\arabic{table}}%
  \renewcommand{\theequation}{A\arabic{equation}}%
  \pagestyle{supplement}%
  \thispagestyle{supplement}%
  \vspace*{1.0em}%
}

\begin{document}

\maketitle

\begin{abstract}
Query denoising has become a standard training strategy for DETR-based detectors.
Denoising queries, initialized by perturbing ground truths, share similarities with track queries in a typical DETR-based Multi-Object Tracking (MOT) method, warranting exploration of their potential synergy.
However, query denoising in existing MOT methods is performed only within a single frame, preventing trackers from learning inter-frame temporal association from the denoising process.
To address this issue, we propose \name, a \underline{T}emporal \underline{Q}uery \underline{D}enoising (TQD) method tailored for MOT. 
In our method, denoising queries are initialized from ground truths in the previous frame and then propagated into the current frame in the same way as track queries, serving as additional independent data association candidates.
These denoising queries carry temporal information and instance-specific feature representations, effectively emulating and augmenting track queries.
Moreover, to simulate various real-world MOT challenges for robust tracking, we introduce several corresponding noise types to generate diverse denoising queries.
We analyze the impact of our temporal query denoising for two tracking paradigms, \textit{tracking-by-attention} and \textit{alternating detection and association}, demonstrating its generalization.
Extensive experiments on the nuScenes \bmvc{and Argoverse~2 datasets} demonstrate that our approach consistently enhances different MOT baselines, requiring only modifications in the training process.
Code and models are available at \ready{https://github.com/yutongy98/TQD-Track}.
\end{abstract}

\section{Introduction}
\label{sec:intro}
Autonomous driving has attracted extensive research attention~\cite{hu2023planning,li2026spacedrive,jiang2023vad,li2025ago,zhang2024seflow,du2026x}, and 3D Multi-Object Tracking (MOT) plays a crucial role in understanding surrounding dynamic agents over time.
In recent years, transformer-based 3D MOT methods~\cite{zhang2022mutr3d, pang2023standing, ding2024ada, li2023end} have been widely adopted due to their seamless integration with DETR-based detectors~\cite{wang2022detr3d, liu2022petr} and end-to-end autonomous driving~\cite{hu2023planning}.
These methods typically leverage track queries to consistently represent the same instances, which can be achieved by combining them with an implicit or explicit data association module.
Despite their promising performance, these methods struggle to deal with rare behaviors of other traffic participants or uncertain scenarios, limiting their robustness and scalability in real-world applications.
Addressing this issue is essential for improving tracking reliability in complex environments.

Beyond 3D MOT, query denoising (DN)~\cite{li2022dn} has become a standard training technique in 2D and 3D DETR-based detectors, where denoising queries are generated by adding noise to ground truth boxes to serve as additional training samples.
Query denoising addresses the slow convergence issue in DETR by separating box refinement from bipartite matching which often yields inconsistent results during early training stages.
Beyond accelerating training, denoising queries provide a convenient way to introduce controlled perturbations and expose the model to diverse training samples without modifying the original data.

\begin{figure}[t!]
    \centering
    \includegraphics[width=\textwidth]{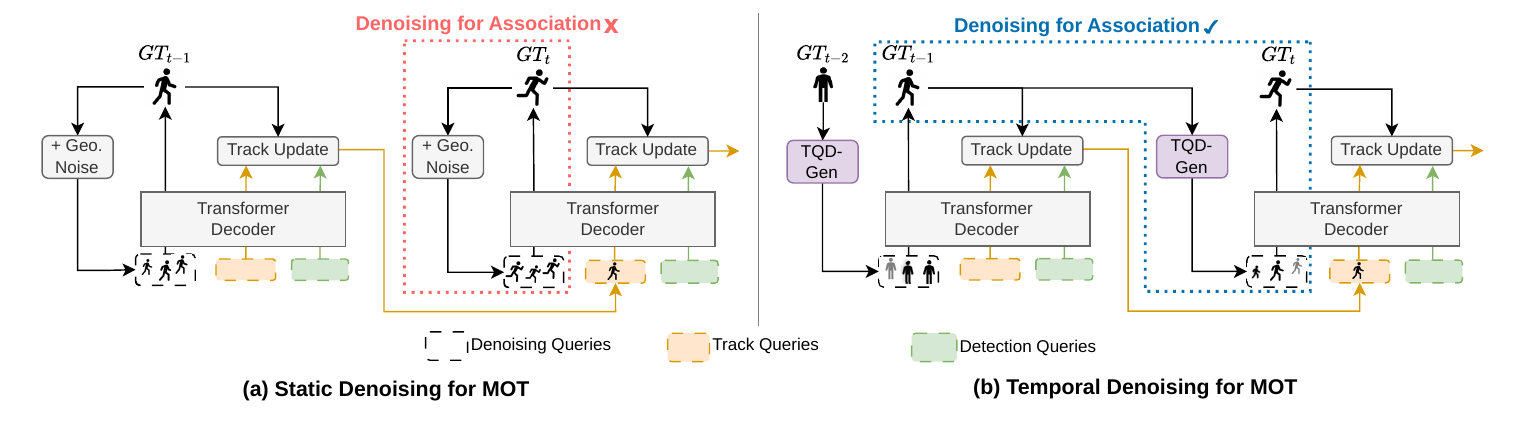}

    \caption{\textbf{Comparison between static and temporal query denoising in MOT.}
    (a) In static denoising, the DN queries are generated from GT boxes at the current frame $t$, failing to contribute to the crucial inter-frame data association in MOT.
    (b) Our approach generates DN queries from GTs at the previous frame $t-1$ using our proposed temporal query denoising generator (TQD-Gen) by adding various types of noise. 
    These DN queries are then propagated to the current frame $t$ for association learning, which aligns with the streaming nature of MOT, thereby effectively enhancing tracking robustness.
    }
    \label{fig:title}\end{figure}

In the context of MOT, both track queries and the aforementioned denoising queries can be regarded as instance-specific representations, each tied to a unique instance and affected by uncertainty and noise.
Consequently, leveraging query denoising to simulate complex and dynamic scenarios in real-world tracking offers a promising way to enhance tracking robustness.
Recent studies~\cite{fu2023denoising, zhou2024ua} 
directly perform denoising query generation and ground truth (GT) reconstruction within the same frame for MOT, a process we call \textit{static query denoising} as illustrated in Fig.~\ref{fig:title}a.
However, unlike object detection, MOT requires the model to operate in a streaming and auto-regressive manner.
Thus, static denoising queries fail to exploit their similarity to track queries to effectively contribute to the MOT task, as they ignore temporal modeling.
Furthermore, since most 3D MOT approaches~\cite{zhang2022mutr3d, pang2023standing, ding2024ada} are initialized from a pre-trained detector, they cannot directly benefit from the accelerated convergence provided by denoising.
These limitations hinder the efficacy of existing denoising strategies for MOT.

To address this challenge, we propose \name{}, a \underline{T}emporal \underline{Q}uery \underline{D}enoising method that explicitly aligns denoising with the auto-regressive tracking process.
As shown in Fig.~\ref{fig:title}b, the denoising queries are generated by adding various noise types to the GT annotations of the previous frame $t-1$ and are then propagated to the current frame $t$.
They are subsequently fed into the tracker as augmented inputs and associated with their corresponding GT boxes in the new frame $t$. 
This enables the denoising queries to effectively simulate real-world tracking scenarios, such as object disappearances and motion uncertainty, to which the track queries must appropriately react in the subsequent frame.
Furthermore, our denoising queries cover various types of noise, including geometric noise, motion noise, query feature noise, and instance-level noise.
This effectively models the inherent uncertainties and self-induced errors (\eg, positional uncertainty and previous false positives) typical of auto-regressive MOT, thereby more accurately reflecting real-world tracking complexities.
We analyze the impact of temporal query denoising on several baseline methods across two different tracking paradigms including tracking-by-attention (TBA)~\cite{zeng2022motr, meinhardt2022trackformer, zhang2022mutr3d} 
and alternating detection and association (ADA)~\cite{ding2024ada}.
For the ADA paradigm with an explicit association module, we further employ an association mask to prevent detection and track queries from aggregating denoising queries, similar to the self-attention masks in~\cite{li2022dn,yang2026groupensemble}.

Extensive experiments on the nuScenes dataset~\cite{caesar2020nuscenes} show that \name{} consistently improves the baseline methods across both the TBA and ADA paradigms.
\bmvc{We further validate its effectiveness on the Argoverse~2 dataset~\cite{wilson2023argoverse} within the ADA paradigm, where TQD-Track delivers consistent performance gains.}
As our method modifies only the training phase, the model architecture and inference time are identical to the original tracker.

\section{Related Work}
\subsection{DETR-based Object Detection}
\ready{DETR~\cite{carion2020end} reformulates object detection as transformer-based set prediction, removing hand-crafted anchors and Non-Maximum Suppression (NMS).
Its variants~\cite{zhu2020deformable, meng2021conditional, liudab, li2022dn, zhangdino} improve convergence and feature sampling, while 3D extensions such as DETR3D~\cite{wang2022detr3d} and PETR~\cite{liu2022petr} enable multi-view 3D detection with query-based spatial reasoning.
These DETR-based detectors provide a flexible foundation for downstream multi-object tracking.}

\subsection{DETR-based Multi-Object Tracking}
MOTR~\cite{zeng2022motr} and TrackFormer~\cite{meinhardt2022trackformer} are pioneering works on DETR-based end-to-end multi-object tracking (MOT).
They regard tracked objects as track queries that are propagated through frames and aim to track the same object instances consistently, while detection (det.) queries initialize newborn tracks.
This establishes the tracking-by-attention (TBA) paradigm which has been further explored in 3D MOT~\cite{zhang2022mutr3d, pang2023standing, zhou2024ua}.
However, recent works~\cite{ding2024ada, wang2024onetrack} point out that TBA struggles in balancing both object detection and tracking tasks.
Beyond TBA, DETR also enables end-to-end MOT under the more traditional tracking-by-detection (TBD) paradigm, where objects are detected using DETR and then associated using heuristic~\cite{xu2022transcenter, zhao2022tracking} or learning-based methods~\cite{li2022time3d,li2023end}.
More recently, ADA-Track~\cite{ding2024ada} proposes an alternating detection and association (ADA) paradigm, which integrates an explicit data association module based on Edge-Augmented Cross-Attention~\cite{Hussain2021EdgeaugmentedGT} into each decoder layer and fully utilizes the synergy between detection and association.
\bmvc{Building on this, ADA-Track++~\cite{ding2024adapp} introduces an auxiliary token to mitigate softmax-induced over-association issue.
Furthermore, SynCL~\cite{lin2026syncl} bridges detection-association gaps via instance-aware contrastive learning, achieving reproducible state-of-the-art results on nuScenes.} 
\ready{
Recent MOT works further enhance inference-time temporal modeling:
OFTrack~\cite{song2025temporal} lifts optical flow to object flow over multi-frame sections,
while HyperSSM~\cite{song2026hypergraph} combines hypergraph reasoning with state-space modeling to improve spatio-temporal motion coherence.
In contrast, TQD-Track improves association learning through auxiliary temporal denoising during training, while leaving inference unchanged.}

\subsection{Query Denoising}
Vanilla DETR~\cite{carion2020end} faces the slow convergence issue because the Hungarian Matching between predicted and GT boxes is inconsistent, resulting in instability in early training stages.
DN-DETR~\cite{li2022dn} introduces the query denoising strategy, which effectively decouples DETR training into bipartite matching and box regression tasks, leading to accelerated training and better performance.
DINO~\cite{zhangdino} improves upon DN-DETR by adding negative samples to the denoising queries.
Query denoising has also become a standard training strategy in various 3D detection methods~\cite{liu2022petr, Wang2023streampetr}.
\bmv{Beyond object detection, SQD-MapNet~\cite{wang2024stream} proposes the Stream Query Denoising (SQD) strategy, which incorporates temporal modeling into the query denoising process to target stationary road elements (\eg, lane boundaries) in HD map reconstruction.}
\ready{DenoisingMOT~\cite{fu2023denoising} introduces query denoising into 2D MOT based on TBA.
However, existing denoising methods for MOT ignore temporal modeling, which is crucial for the tracking task.
To address this gap, we propose \name{}.

Compared with DenoisingMOT~\cite{fu2023denoising} and SQD-MapNet~\cite{wang2024stream}, the key distinctions of \name{} are three-fold: (1)~\name{} generates DN queries from frame $t{-}1$, propagates them to frame $t$, and directly supervises cross-frame association; (2)~it introduces MOT-specific center, velocity, query-feature, and instance-level noises to simulate various tracking uncertainties; and (3)~hybrid grouping enables both targeted correction and robustness to mixed uncertainties.
These design choices lead to consistent tracking gains across our experiments.}

\section{Proposed Method}
\begin{figure*}[t!]
      \centering
      \includegraphics[width=\textwidth]{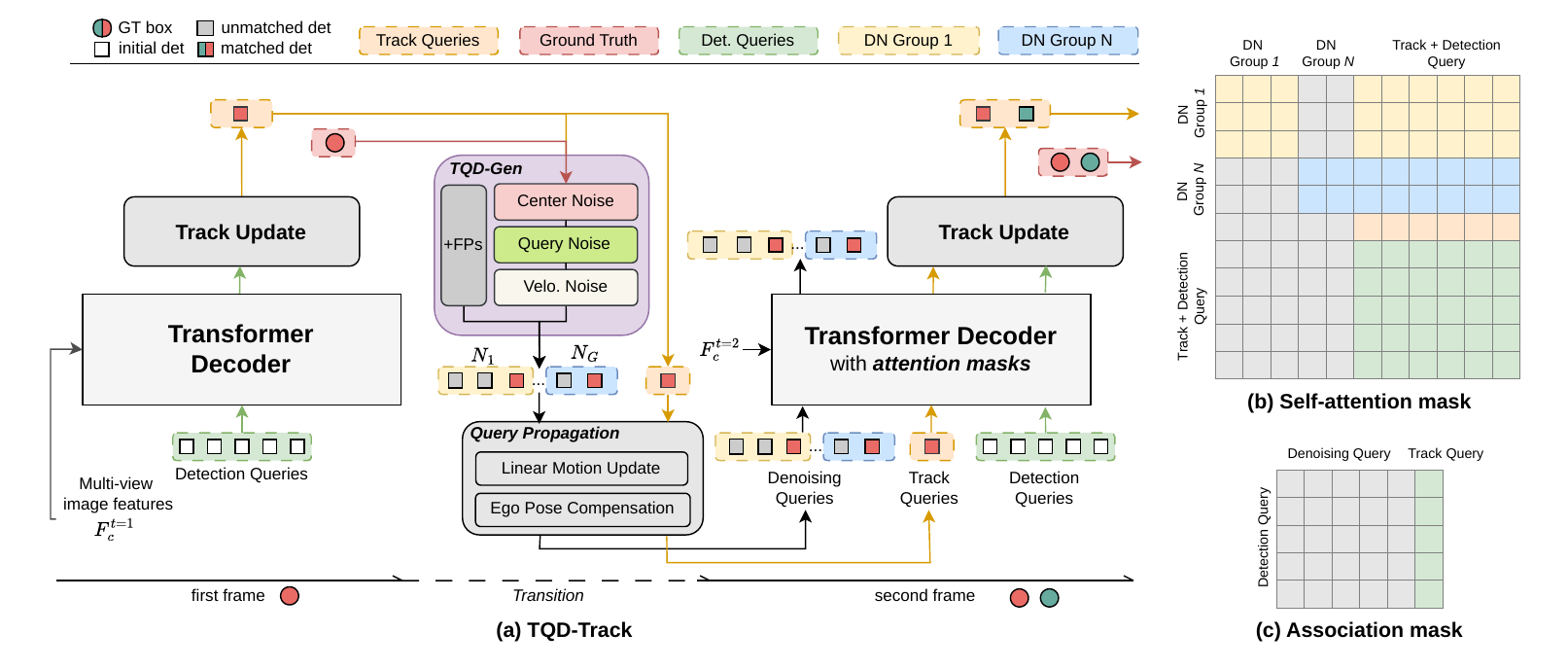}
      \caption{
        \textbf{Overview of \name{} applied to a DETR-based tracker. }
        For a frame $t-1$ (\eg, the first frame in the figure), we generate several ($N_\text{G}$) groups of denoising queries within the Temporal Query Denoising Generator (TQD-Gen) by introducing diverse noise types to the ground truth.
        The denoising queries are propagated to frame $t$ and integrated into the decoder as additional independent association samples.
        The decoder then associates them with their corresponding ground truth instances at the current frame $t$, resulting in the denoising loss $\mathscr{L}_\text{DN}$.
        To ensure alignment between training and inference, we employ a self-attention mask (b) and/or an association mask (c), depending on the specific tracker baseline.
        Gray grids in (b) and (c) denote blocked attention.
        During inference, TQD-Gen is deactivated, thereby preserving the original inference time and model complexity.
      }
      \label{fig:overview}
  \end{figure*}

The overview of the proposed \name{} is shown in Fig. \ref{fig:overview}.
During training, $N_\text{G}$ groups of denoising queries are created from the ground truths of the previous frame $t-1$ by the Temporal Query Denoising Generator (TQD-Gen).
Each denoising group incorporates various types of noise, including \textit{geometric noise}, \textit{query noise}, \textit{motion noise}, and \textit{instance-level noise}.
Similar to track queries, the denoising queries are propagated from frame $t-1$ to $t$, aggregate features from detection queries layer-by-layer, and regress their corresponding ground-truth boxes at $t$.
At each decoder layer, attention masks are applied to prevent leakage of ground truth information from denoising queries to track and detection queries, and ensure an identical query feature forward pass during training and inference.
As query denoising is a training strategy, the model architecture and inference process are identical to the original tracker, maintaining the same inference speed and model complexity.
 As a training-time method, \name{} can be seamlessly integrated into various query-based tracking paradigms, including \textit{Tracking-by-Attention} (TBA) and \textit{Alternating Detection and Association} (ADA).

\subsection{Temporal Query Denoising for MOT}
\label{subsec:tdmot}

\subsubsection{Temporal Query Denoising Generator} The Temporal Query Denoising Generator (TQD-Gen) generates denoising queries $Q_\text{DN}^{t-1}$ across two consecutive frames, from $t-1$ to $t$.
TQD-Gen takes as input the GT boxes $B_\text{GT}^{t-1}$, the track queries $\hat{Q}_\text{T}^{t-1}$, as well as the false positive (FP) detections $\left\{Q_\text{FP}^{t-1}, C_\text{FP}^{t-1} \right\}$, and outputs the denoising queries $Q_\text{DN}^{t-1}$ and their corresponding 3D reference points $C_\text{DN}^{t-1}$:

\begin{equation}
    Q_\text{DN}^{t-1}, C_\text{DN}^{t-1} = \text{TQD-Gen} \left( B_\text{GT}^{t-1}, \hat{Q}_\text{T}^{t-1}, Q_\text{FP}^{t-1}, C_\text{FP}^{t-1} \right)
\end{equation}
Each GT box $b_{\text{GT},i}^{t-1}=\{c_i, s_i, \theta_i, v_i\} \in B_\text{GT}^{t-1}$ is described by its box center (also the reference point) $c_i \in \mathbb{R}^3$, box size $s_i \in \mathbb{R}^3$, yaw $\theta_i \in \mathbb{R}$, and the velocities $v_i \in \mathbb{R}^2$ in the Bird's Eye View (BEV) plane.
The bounding boxes of the denoising queries $B_\text{DN}^{t-1}$ are initialized from $B_\text{GT}^{t-1}$ and inherit their GT instance IDs.
For each denoising box $b_{\text{DN},i}^{t-1}$, its corresponding denoising query features are copied from the track query of the same instance ID, e.g. $q_{\text{DN},i}^{t-1} = q_{\text{T},j}^{t-1}$ with $\text{ID}(b_{\text{DN},i}^{t-1}) = \text{ID}(q_{\text{T},j}^{t-1})$.
We apply the following types of noise in the TQD-Gen:

\paragraph{Center noise}
For the denoising reference point $c_{\text{DN},i}^{t-1} \in C_\text{DN}^{t-1}$ of $i$-th query, we first add uniformly distributed noise $(\Delta x_i, \Delta y_i, \Delta z_i) $ to the 3D GT box center $c_i$:
\begin{equation}
    \label{equ:centernoise}
    c_{\text{DN},i}^{t-1} \leftarrow c_{\text{DN},i}^{t-1} + \left[ \Delta x_i, \Delta y_i, \Delta z_i \right]^{T}
\end{equation}
A hyper-parameter $\lambda$ is used to control the sampling range of noise which scales linearly to the object size $s_i=\{w_i,l_i,h_i\}$, \ie, $\left| \Delta x_i \right|\lt \frac{\lambda w_i}{2}$, $ \left| \Delta y_i \right|\lt \frac{\lambda l_i}{2}$ and $\left| \Delta z_i \right|\lt \frac{\lambda h_i}{2}$. 

\paragraph{Query noise}
For the denoising query $q_{\text{DN},i}^{t-1}$, we add Gaussian noise $\boldsymbol{\epsilon}_i^\text{query} \sim \mathscr{N}(\boldsymbol{0}, \boldsymbol{\Sigma}^\text{query})$ onto the query feature:
\begin{equation}
    \label{eq:appearancenoise}
    q_{\text{DN},i}^{t-1} \leftarrow q_{\text{DN},i}^{t-1} + \boldsymbol{\epsilon}_i^\text{query},
\end{equation}
where $\boldsymbol{\epsilon}_i^\text{query}$ has the same dimension as $q_{\text{DN},i}^{t-1}$.
The purpose of the query noise is to enhance the model's resilience to feature variations by ``blurring'' the feature representation.

\paragraph{Velocity noise}
As the denoising queries will be propagated to the next frame via a linear motion model, we also add Gaussian noise $\boldsymbol{\epsilon}_i^\text{velo} \sim \mathscr{N}(\boldsymbol{0}, \boldsymbol{\Sigma}^{velo})$ to the velocities $v_{\text{DN},i}$ to model the motion noise in the transition phase, i.e.,
\begin{equation}
    \label{eq:velonoise}
    v_{\text{DN},i} \leftarrow v_{\text{DN},i} + \boldsymbol{\epsilon}_i^\text{velo}
\end{equation}
This velocity noise simulates the uncertainty in the object's motion, providing the tracker with diverse samples during training to enhance robustness.
\bmvc{The incorporation of velocity and query noise distinguishes our method from existing static denoising methods and SQD-MapNet~\cite{wang2024stream}, both of which ignore motion-related noise.}

\begin{figure}[t!]
    \centering
    \includegraphics[width=\linewidth]{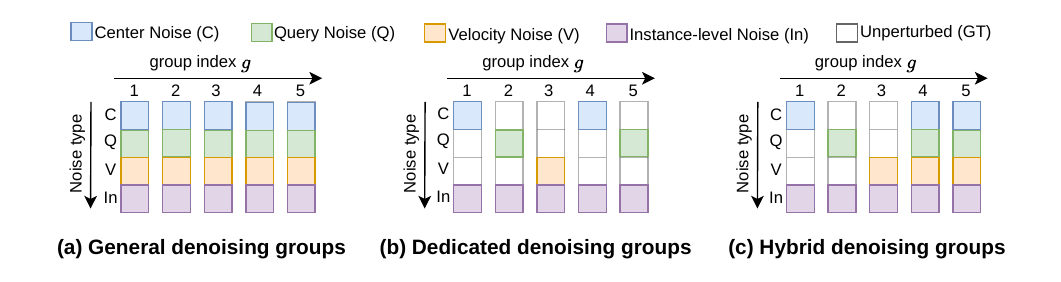}
    \caption{\textbf{Illustration of different denoising group strategies.} 
    In this example, we illustrate the concept using 5 denoising groups.
    (a) General denoising groups incorporate all available noise types within every group.
    (b) Dedicated denoising groups assign a specific noise type (e.g., center, query, or velocity noise) to each group, built upon the fundamental instance-level noise. 
    White cells indicate that the corresponding attributes retain their unperturbed ground-truth values. 
    (c) Hybrid denoising groups combine both strategies to comprehensively enhance the tracker's robustness against diverse forms of uncertainties.}
    \label{fig:denoising_groups}
\end{figure}

\paragraph{Instance-level noise}
To \bmvc{model self-induced errors that are inherent to the auto-regressive nature of MOT, we introduce negative samples into the denoising queries, helping the model correctly rectify errors propagated from previous frames by suppressing false positive track queries.}
We achieve this by sampling false positive detection queries that fail to match with any ground truth instances \bmvc{during the training phase} from $\left\{Q_\text{FP}^{t-1}, C_\text{FP}^{t-1} \right\}$.
Concretely, the FP detections with the top $k$ confidence scores will be concatenated to the denoising queries, \ie,
    \begin{equation}
    \label{eq:addfp}
    Q_\text{DN}^{t-1} \leftarrow [Q_\text{DN}^{t-1} \ \| \ \text{TopK} (Q_\text{FP}^{t-1})],
    \ C_\text{DN}^{t-1} \leftarrow [C_\text{DN}^{t-1} \ \| \ \text{TopK} (C_\text{FP}^{t-1})],
\end{equation}
where $k = \alpha^{FP} N_\text{GT}^{t-1}$, with $\alpha^\text{FP}$ regulating the ratio between the number of false positive samples and the number of GT instances $N_\text{GT}^{t-1}$.
$[\ \cdot \ \| \ \cdot \ ]$ denotes concatenation.

\subsubsection{Denoising Groups}
\bmvc{We repeat the aforementioned process $N_\text{G}$ times to generate multiple denoising groups, producing a global set of denoising queries $\left\{ Q_\text{DN}^{t-1,g}, C_\text{DN}^{t-1,g} \right\}_g$ where $g \in \{1, \dots, N_G\}$.
The noise sampling for each group is performed independently.
In standard object detection frameworks~\cite{li2022dn,zhangdino}, each denoising group is formed by incorporating all available noise types without distinction, which we refer to as general denoising groups (Fig.~\ref{fig:denoising_groups}a).
However, unlike static object detection, the streaming nature of MOT task introduces intricate temporal uncertainties.
While providing broad noise exposure, general groups fail to disentangle the individual impacts of mixed uncertainties, preventing the model from learning targeted error correction effectively.
To this end, we introduce target-specific denoising groups, termed \textit{dedicated} denoising groups (Fig.~\ref{fig:denoising_groups}b).
In this strategy, each group is assigned a single specific noise type (\eg, center, query, or velocity noise) alongside the fundamental instance-level noise.
Crucially, unassigned attributes within a group retain their unperturbed ground-truth values.
This decoupling aims to sharpen the model's ability to resolve specific error types.
In addition, \textit{dedicated} denoising groups can be combined with \textit{general} denoising groups, resulting in \textit{hybrid} denoising groups (Fig.~\ref{fig:denoising_groups}c), leading to a wide spectrum of uncertainties.}

\subsubsection{Query Propagation}
\label{subsec:query_p}
Once the denoising queries are generated at $t-1$, they will be propagated to the next frame $t$ \bmvc{via a linear motion update and ego pose compensation (Fig. \ref{fig:overview}a), along with the track queries.}
\bmvc{A key advantage of our proposed temporal query denoising lies in its ability to handle variations in GT instances across adjacent frames, such as object disappearances during frame transitions. 
These situations can be effectively simulated in temporal query denoising by propagating denoising queries across time steps, closely aligning with the inherent behavior of track queries.}

\subsubsection{Denoising Process}
In standard static query denoising~\cite{li2022dn}, denoising queries aim to reconstruct their corresponding GT boxes
within the same frame, aligning with the detection task.
In the DETR-based MOT method, the track queries are tasked with detecting their corresponding GT instances at $t$ that they have already detected at previous frame $t-1$.
Similarly, in our temporal query denoising method, denoising queries are generated in frame $t-1$ and subsequently associated with their corresponding GT boxes in the current frame $t$, highly aligned with the streaming nature of MOT.
These denoising queries are fed into the model as augmented track queries and participate in attending to image features and interacting with detection queries together with the attention masks (see the following subsection).
\bmv{For training, the denoising loss term $\mathscr{L}_\text{DN}$ is computed by supervising the DN query predictions with their corresponding GT boxes at frame $t$, encompassing classification loss (focal loss) and box regression losses ( $L_1$ and GIoU losses), mirroring the standard training objective of the track queries.
While positive DN queries learn to reconstruct their corresponding GT boxes, those generated from previous-frame false positives, \ie, negative DN samples, are supervised as background, helping the tracker suppress its own previous tracking errors during inference.}
This results in the overall loss $\mathscr{L} = \lambda_\text{DN}\mathscr{L}_\text{DN} + \mathscr{L}_\text{tracker}$, where $\mathscr{L}_\text{tracker}$ is the loss of the selected base tracker.
As a training strategy, the DN queries are only generated during training, which ensures the same inference time and model complexity as the baseline tracker.

\subsection{Attention Masks}
\label{subsec:masks}
Attention masks are essential in query denoising~\cite{li2022dn} to ensure that interactions between detection and track queries remain consistent with those during inference.
Depending on the tracking paradigms, different masking strategies are employed.
Specifically, TBA requires only a self-attention mask, while ADA needs an additional association mask for its data association module.

\subsubsection{Self-attention Mask}
In self-attention, we prevent the track and detection queries from attending to the denoising queries while allowing attention in the opposite direction.
In addition, attention between denoising queries in different groups is blocked, ensuring the independence of each denoising group.
\ready{Formally, let $N_\text{DN,all}=\sum_{g=1}^{N_\text{G}}N_\text{DN}^g$ and $W=N_\text{DN,all}+N_\text{T}+N_\text{D}$.
The queries are concatenated in the order of denoising, track, and detection queries.
For a query index $i$, we define its denoising group indicator as:}
\begin{equation}
\ready{
    \mathrm{grp}(i)=
    \begin{cases}
        g, & \text{if } \sum_{h=1}^{g-1}N_\text{DN}^h < i \le \sum_{h=1}^{g}N_\text{DN}^h,\ g\in\{1,\ldots,N_\text{G}\},\\
        0, & \text{if } i>N_\text{DN,all},
    \end{cases}}
\end{equation}
\ready{where $\mathrm{grp}(i)=0$ denotes a track or detection query.
The self-attention mask $M_\text{SA}=[m^\text{SA}_{ij}]_{W\times W}$ is defined as:}
\begin{equation}
\ready{
    m^\text{SA}_{ij} =
    \begin{cases}
        1, & j\le N_\text{DN,all} \text{ and } \mathrm{grp}(i)\neq \mathrm{grp}(j),\\
        0, & \text{otherwise}.
    \end{cases}}
\end{equation}
\ready{Here, $m^{\mathrm{SA}}_{ij}=1$ means that query $i$ cannot attend to query $j$. 
Thus, DN groups are mutually isolated, and normal track/detection queries cannot attend to any DN query.
Fig. \ref{fig:overview}b illustrates the self-attention mask, where the gray grids denote the blocked attention.
}

\subsubsection{Association Mask}
\bmv{
For the explicit data association module in ADA~\cite{ding2024ada}, we treat denoising queries as augmented track queries within a unified set $Q_\text{U} = \bigl\{ Q_{\text{DN}}, \ Q_\text{T} \bigr\}$.
To prevent ground truth (GT) information leakage from denoising to detection queries while maintaining association-related feature interaction, we introduce an association mask (Fig.~\ref{fig:overview}c).
This allows denoising queries to effectively contribute to association learning while ensuring the alignment between the training and inference phases.
Detailed mathematical formulations are in the supplement.}

\section{Experiments}

\subsection{Experiment Setup}
\label{subset: setup}

\subsubsection{Dataset}
\bmvc{
We evaluate our method on the nuScenes~\cite{caesar2020nuscenes} and Argoverse 2 (AV2)~\cite{wilson2023argoverse} datasets, each featuring 1,000 scenes with 6- and 7-camera ring setups, respectively. While nuScenes provides 20s clips annotated at 2 Hz, AV2 offers 15s clips annotated at 10 Hz. In all experiments, we use multi-camera images as the sole input.

\noindent \textbf{Category Selection.} We focus on dynamic objects relevant to 3D MOT. 
For nuScenes, we follow the standard protocol by using the seven primary dynamic categories.
\bmv{For AV2, we filter its original 26 fine-grained classes by removing eight static or irrelevant ones (e.g., signs, animals). 
The remaining 18 classes are mapped into nine classes by merging those with diverse sub-types (e.g., merging various bus types into \textit{Bus}) while preserving distinct standalone classes such as \textit{Regular Vehicle} and \textit{Pedestrian}.
These nine categories are used for both training and evaluation.
Detailed class mapping is provided in the supplement.}}

\subsubsection{Evaluation Metrics}
\bmvc{For the nuScenes dataset}, we evaluate \name{} on its official primary MOT metrics, including Average Multi-Object Tracking Accuracy (AMOTA) and Average Multi-Object Tracking Precision (AMOTP)~\cite{Weng20193DMT}, as well as secondary metrics from CLEAR MOT~\cite{bernardin2008evaluating} such as MOTA, MOTP, ID-switches (IDS), and Recall.
\bmvc{
For the Argoverse 2 dataset, we report Higher-Order Tracking Accuracy (HOTA)~\cite{luiten2021hota}, the official primary metric for this benchmark, following the standard \SI{50}{m} ego-distance threshold.}

\subsubsection{Implementation Details}
\bmvc{
\noindent \textbf{Training Details.} We adopt MUTR3D~\cite{zhang2022mutr3d} and ADA-Track~\cite{ding2024ada} as baselines on nuScenes, configured with DETR3D~\cite{wang2022detr3d} and PETR~\cite{liu2022petr} detectors.
For AV2, DETR3D-based ADA-Track is used as the primary baseline.
VoVNetV2~\cite{lee2019energy} is adopted as the detector backbone for all experiments.
Trackers are initialized with detector weights pre-trained over 24 epochs, consistent with existing works~\cite{zhang2022mutr3d, pang2023standing, ding2024ada}.
Following baseline protocols, each training sample is a 1.5s snippet (3 frames) for nuScenes.
To maintain this temporal window, we downsample AV2 from 10 Hz to 2 Hz, aligning with common practice in the recent AV2 benchmark~\cite{chenle3de2e}.
We train for 24 epochs on 8 A100 GPUs for nuScenes, and 18 epochs on 4 A100 GPUs for AV2 for computation efficiency, both with a batch size of 1 per GPU.

\bmv{
\noindent \textbf{Static Denoising.} 
For a fair comparison, static DN shares the same tracker architecture and hyperparameters as TQD. 
It differs only in denoising initialization (Fig.~\ref{fig:title}): static DN uses current-frame ground truths (GTs) with center noise following~\cite{fu2023denoising,li2022dn,zhangdino}, while TQD uses previous-frame GTs with TQD-Gen and DN query propagation. 
Both are trained against current-frame GTs, isolating the impact of temporal modeling in the denoising process.
}

\noindent \textbf{Hyperparameters.}  
Based on the ablation studies on nuScenes, we set the uniform center noise scale $\lambda=1$;
the velocity noise covariance $\boldsymbol{\Sigma}^{velo} = \sigma_{velo}\mathbf{I} \in \mathbb{R}^{2 \times 2}$ where $\sigma_{velo}=4$ and $\mathbf{I}$ is the identity matrix;
the query feature noise covariance $\boldsymbol{\Sigma}^{query} = \sigma_{query}\mathbf{I} \in \mathbb{R}^{d_k \times d_k}$ where $\sigma_{query}=0.1$ and $d_k=256$;
the false positive ratio $\alpha^{FP}=0.1$;
the number of denoising groups $N_{\text{G}}=5$ using the hybrid grouping strategy. 
The denoising loss weight $\lambda_\text{DN}$ is set to~1.
We adopt these same hyperparameters for AV2 and observe consistent performance gains.}

\begin{table}[t!]
    \centering
    
    \stepcounter{table} 
    
    \subtable[\textbf{Performance comparison on the nuScenes validation set.}
    ]{
        \resizebox{\textwidth}{!}{ 
            \setlength{\tabcolsep}{1.0mm} 
            \begin{tabular}{c|c|c|ccccc}
                \toprule
                Paradigm & Detector & Denoising Method & AMOTA$\uparrow$ & AMOTP$\downarrow$ & Recall$\uparrow$ & MOTA$\uparrow$ & IDS$\downarrow$ \\
                \midrule
                \multirow{6}{*}{ \shortstack{MUTR3D~\cite{zhang2022mutr3d} \\ (\textit{TBA baseline})} } 
                      & \multirow{3}{*}{DETR3D} & w/o   &   0.362    &    1.375   &   0.479    &   0.335    &    641   \\
                      &        & Static &   0.368$^{(\uparrow 0.6\%pt)}$    &   1.386    &   0.469   &   0.340    &   652   \\
                      &        & Temporal (Ours) &   \textbf{0.387}$^{(\uparrow \textbf{2.5}\%pt)}$    &   \textbf{1.359}    &   \textbf{0.505}    &   \textbf{0.351}    &   \textbf{637}  \\
                      \cmidrule{2-8}          
                      & \multirow{3}{*}{PETR} & w/o   & 0.443 & 1.299 & 0.552 & 0.416 & 175 \\
                      &        & Static & 0.445$^{(\uparrow 0.2\%pt)}$ & 1.303 & 0.541 & 0.422 & \textbf{152} \\
                      &        & Temporal (Ours) & \textbf{0.460}$^{(\uparrow \textbf{1.7}\%pt)}$ & \textbf{1.281} & \textbf{0.565} & \textbf{0.441} & 160 \\
                \midrule
                \multirow{6}{*}{ \shortstack{ADA-Track~\cite{ding2024ada} \\ (\textit{ADA baseline})} } 
                      & \multirow{3}{*}{DETR3D} & w/o   & 0.420  & 1.320  & \textbf{0.543} & 0.389 & 795  \\
                      &        & Static & 0.422$^{(\uparrow 0.2\%pt)}$ & \textbf{1.310} & 0.537 & 0.388 & 848   \\
                      &        & Temporal (Ours) & \textbf{0.441}$^{(\uparrow \textbf{2.1}\%pt)}$ & \textbf{1.310} & \textbf{0.543} & \textbf{0.408} & \textbf{752} \\
                      \cmidrule{2-8}          
                      & \multirow{3}{*}{PETR} & w/o   & 0.493 & 1.208 & 0.602 & 0.450  & 672 \\
                      &        & Static & 0.496$^{(\uparrow 0.3\%pt)}$ & 1.216 & 0.604 & 0.453 & 632 \\
                      &        & Temporal (Ours) & \textbf{0.515}$^{(\uparrow \textbf{2.2}\%pt)}$ & \textbf{1.206} & \textbf{0.623} & \textbf{0.467} & \textbf{596} \\
                \bottomrule
            \end{tabular}
        }
        \label{tab:dncom}
    }

    \subtable[\textbf{Performance comparison on the Argoverse 2 validation set.} 
    ]{
        \resizebox{\textwidth}{!}{ 
            \begin{tabular}{l|ccccccccc|c} 
            \toprule
            Method & Reg. Veh. & Ped. & Bicy. & Trai. & Moto. & Rider & Bus & Pers. Mob. & Heav. Veh. & Avg. HOTA$\uparrow$ \\
            \midrule
            ADA-Track\cite{ding2024ada} & 0.359 & 0.239 & 0.232 & 0.177 & 0.293 & 0.259 & 0.283 & 0.191 & 0.241 & 0.253 \\
            + Static DN & 0.368 & 0.239 & 0.241 & 0.182 & 0.294 & 0.244 & 0.282 & 0.189 & 0.243 & 0.254 \\
            \textbf{+ Temp. DN} & \textbf{0.379} & \textbf{0.246} & \textbf{0.247} & \textbf{0.191} & \textbf{0.298} & \textbf{0.262} & \textbf{0.288} & \textbf{0.202} & \textbf{0.249} & \textbf{0.262} \\
            \bottomrule
            \end{tabular}
        }
        \label{tab:per_class_hota}
    }

    \addtocounter{table}{-1} 
    
    \caption{\textbf{Effectiveness and Generalizability of Temporal Query Denoising.} \textbf{(a)} On the primary nuScenes benchmark, our method consistently outperforms both vanilla training and static denoising, demonstrating its robust adaptability to different tracking paradigms and detectors. 
    \textbf{(b)} On the Argoverse 2 dataset, the steady performance gains across all object categories further support the effectiveness and generalizability of our training strategy.
    \textit{Class abbreviations}: Reg. Veh. (Regular Vehicle), Ped. (Pedestrian), Bicy. (Bicycle), Trai. (Trailer), Moto. (Motorcycle), Pers. Mob. (Personal Mobility), Heav. Veh. (Heavy Vehicle).
    }
    \label{tab:merged_main_results}
\end{table}

\subsection{Experimental Results}
\label{subset: dn_compare}
\subsubsection{Query Denoising Strategy Comparison}
To demonstrate the effectiveness of our method, we compare our proposed Temporal Query Denoising with Static Query Denoising (Fig.~\ref{fig:title}a) and the vanilla training strategy in Table~\ref{tab:merged_main_results}.

\bmvc{For the nuScenes dataset (Table~\ref{tab:dncom})}, across both TBA and ADA paradigms, a consistent performance pattern emerges. 
Static denoising leads to only marginal AMOTA gains (less than 0.6\%pt) for either detector, reflecting its inherently limited impact as it operates solely within a single frame and cannot contribute to temporal association learning. 
In contrast, our temporal denoising delivers substantial and consistent improvements, raising AMOTA by +2.5\%pt and +1.7\%pt for TBA, and +2.1\%pt and +2.2\%pt for ADA with DETR3D and PETR, respectively.
The consistent improvements over static denoising highlight the generalization capability of temporal denoising to different tracking architectures.
These gains can be attributed to the better alignment of temporal denoising with the auto-regressive inference scheme in MOT, enabling more effective training sample augmentation.
In addition, diverse temporally targeted noise types introduce rare motion patterns and challenging association cases throughout training, thereby strengthening tracker robustness.

\bmvc{Notably, similar performance trends are observed on the Argoverse 2 dataset (Table~\ref{tab:per_class_hota}).
Despite the significant disparities in sensor configurations and environmental distributions between the two benchmarks, TQD-Track consistently outperforms the ADA-Track baseline, achieving 0.262 Avg. HOTA (+0.9\%pt over the baseline and +0.8\%pt over static DN), demonstrating its robust generalizability.
Crucially, the per-class analysis validates our motivation of addressing the inherent challenges of dynamic object association in MOT with temporal denoising.
We observe significant benefits in categories with highly variable kinematic characteristics, such as Regular Vehicle (+2.0\%pt), Bicycle (+1.5\%pt), and Trailer (+1.4\%pt).
In contrast, static denoising yields limited gains on these classes due to its inability to contribute to crucial association learning.}
\bmv{
Consistent trends are also reflected in nuScenes, where TQD-Track excels in categories exhibiting complex motion (\eg, Motorcycle +2.3\%pt, Bicycle +2.2\%pt) or challenging articulation (e.g., Trailer +9.1\%pt).
Detailed per-class results and analysis for nuScenes are provided in the supplementary material.

These gains collectively show that the MOT-aligned temporal denoising and diverse noise types in TQD-Track effectively augment the training distribution to better reflect real-world MOT complexities, enhancing tracking robustness for dynamic objects. }

\begin{table*}[t!]
  \centering
  \resizebox{\linewidth}{!}{
    \begin{tabular}{c|c|c|c|ccccc}
    \toprule
    Method & Reference & Detector & Backbone & AMOTA$\uparrow$ & AMOTP$\downarrow$ & Recall$\uparrow$ & MOTA$\uparrow$ & IDS$\downarrow$ \\
    \midrule
    MUTR3D~\cite{zhang2022mutr3d} & CVPRW'22 & DETR3D & R101  & 0.294 & 1.498 & 0.427 & 0.267 & 3822 \\
    DQTrack~\cite{li2023end} & ICCV'23 & DETR3D & R101  & 0.367 & 1.351 & --    & --    & 1120 \\
    STAR-Track~\cite{doll2023star} & RA-L'23 & DETR3D & R101  & 0.379 & 1.358 & 0.501 & 0.360  & \textbf{372} \\
    ADA-Track~\cite{ding2024ada} & CVPR'24 & DETR3D & V2-99 & 0.423 & \textbf{1.298} & \textbf{0.550} & 0.391 & 732 \\
    ADA-Track++~\cite{ding2024adapp} & TPAMI'25 & DETR3D  & V2-99 & 0.434 & 1.304 & 0.556 & 0.386 & 863 \\
    \textbf{TQD-Track (Ours)} & - & DETR3D & V2-99 & \textbf{0.441} & 1.310  & 0.543 & \textbf{0.408} & 752 \\
    \midrule
    MUTR3D$^*$~\cite{zhang2022mutr3d} & CVPRW'22 & PETR  & V2-99 & 0.443 & 1.299 & 0.552 & 0.416 & \textbf{175} \\
    PF-Track~\cite{pang2023standing} & CVPR'23 & PETR  & V2-99 & 0.479 & 1.227 & 0.590  & 0.435 & 181 \\
    ADA-Track~\cite{ding2024ada} & CVPR'24 & PETR  & V2-99 & 0.493 & 1.208 & 0.602 & 0.450  & 672 \\
    ADA-Track++~\cite{ding2024adapp} & TPAMI'25 & PETR  & V2-99 & 0.504 & 1.197 & 0.608 & 0.445 & 613 \\
    SynCL~\cite{lin2026syncl} & NeurIPS'25 & PETR  & V2-99 & 0.507 & \textbf{1.183} & 0.613 & 0.462 & 248 \\
    \textbf{TQD-Track (Ours)} & - & PETR  & V2-99 & \textbf{0.515} & 1.206 & \textbf{0.623} & \textbf{0.467} & 596 \\
    \bottomrule
    \end{tabular}%
  }
  \caption{\textbf{Comparison of \name{} with existing methods on the nuScenes validation set.} $^*$ denotes reproduced results.
  }
  \label{tab:sotacom}%
\end{table*}%

\subsubsection{Comparison with State-of-the-Art on nuScenes}
Table~\ref{tab:sotacom} reports the comparison of \name{} with representative transformer-based multi-camera 3D MOT methods on the nuScenes validation set. 
For the DETR3D detector, \name{} achieves the best AMOTA (0.441) and MOTA (0.408) among all methods, outperforming the previous best method, \bmvc{ADA-Track++, by 0.7\%pt AMOTA and a significant 2.2\%pt MOTA}
while maintaining competitive AMOTP (1.310) and Recall (0.543). Compared to earlier TBA-style methods such as MUTR3D and STAR-Track, \name{} shows much larger margins (+14.7\%pt and +6.2\%pt AMOTA, respectively), demonstrating the effectiveness of the proposed temporal denoising for association learning.
For the PETR detector, a similar trend is observed. \name{} achieves the highest AMOTA (0.515) and MOTA (0.467), surpassing
\bmvc{the current reproducible state-of-the-art method, SynCL, by +0.8\%pt AMOTA and +0.5\%pt MOTA.}
It also improves Recall to 0.623, indicating enhanced detection and tracking completeness, 
\bmvc{while maintaining a competitive AMOTP (1.206).}

\subsection{Ablation Study}
\label{subset: ablation}
We conduct ablation studies on denoising groups, denoising targets, and denoising query composition.
\bmvc{All ablations use default PETR-based TQD-Track settings on the nuScenes validation set.}
Additional ablation studies are discussed in the supplementary material.

\subsubsection{Denoising Groups}
Table \ref{tab:dn_groups} shows the ablation study on the denoising group strategies, \textit{General}, \textit{Dedicated}, and \textit{Hybrid}, and the number of denoising groups $N_\text{G} \in \{1, 3, 5\}$.
All denoising groups include instance-level noise.
When only one \textit{general} denoising group is applied, \ie, all noise types are used in this group, the tracking performance is similar to the baseline.
This is outperformed by $N_\text{G}=3$ denoising groups by \SI{0.6}{\%pt} in AMOTA, showing the importance of multiple denoising groups to augment the diversity of training samples.
With $N_\text{G}=3$ denoising groups, we can use the \textit{dedicated} grouping by injecting only one of the three noise types (center,  velocity, and query noise) per group.
This results in an \SI{0.4}{\%pt} AMOTA increase compared to \textit{general} grouping, highlighting the effectiveness of having \textit{dedicated} noise groups to reduce interference between noise types.
Interestingly, increasing the group number $N_\text{G}$ from three to five does not yield a further increase in the AMOTA for \textit{general} grouping.
However, by adopting our \textit{hybrid} grouping strategy we can continue to improve the performance using more denoising groups.
To form the \textit{hybrid} group, we combine the three \textit{dedicated} groups with two additional \textit{general} groups.
This configuration outperforms both, the model trained with $N_\text{G}=3$ \textit{dedicated} groups and the model with $N_\text{G}=5$ \textit{general} groups, by \SI{1.2}{\%pt} and \SI{1.7}{\%pt} AMOTA, respectively.
It also achieves a notably higher Recall.
\bmvc{This \textit{hybrid} strategy maximizes denoising efficiency by simultaneously learning targeted corrections for specific noise types and robustness against compounded uncertainties.}
Consequently, we use 5 denoising groups with \textit{hybrid} grouping as our default setting.

\subsubsection{Denoising Targets}
\begin{table}[t!]
    \centering
    \stepcounter{table}
    \footnotesize 
    \setlength{\tabcolsep}{0.6mm} 
    
    \begin{minipage}{0.49\textwidth}
        \centering
        \subtable[\textbf{Denoising groups.}]{
            \begin{tabular}{c|c|ccccc}
            \toprule
            $N_\text{G}$ & Strat.  & AMOTA$\uparrow$ & AMOTP$\downarrow$ & Recall$\uparrow$ & MOTA$\uparrow$ & IDS$\downarrow$ \\
            \midrule
            0     & --    & 0.493 & 1.208 & 0.602 & 0.450  & 672 \\
            \midrule
            1     & Gen. & 0.493 & 1.207 & 0.608 & 0.449 & \textbf{573} \\
            \midrule
            \multirow{2}[2]{*}{3} & Gen. & 0.499 & 1.208 & 0.607 & 0.456 & 600 \\
                  & Ded. & 0.503 & 1.202 & 0.614 & 0.461 & 592 \\
            \midrule
            \multirow{2}[2]{*}{5} & Gen. & 0.498 & \textbf{1.197} & 0.606 & 0.446 & 643 \\
                  & Hyb. & \textbf{0.515} & 1.206 & \textbf{0.623} & \textbf{0.467} & 596 \\
            \bottomrule
            \end{tabular}
            \label{tab:dn_groups}
        }
    \end{minipage}
    \hfill 
    \begin{minipage}{0.49\textwidth}
        \centering
        \subtable[\textbf{Denoising targets.}]{ 
            \begin{tabular}{c|cccc|cc}
                \toprule
                \multirow{2}[2]{*}{Exp.} & Geo. & \multicolumn{2}{c}{Motion} & Inst. & \multirow{2}[2]{*}{AMOTA$\uparrow$} & \multirow{2}[2]{*}{AMOTP$\downarrow$} \\
                \cmidrule{2-5}          & Ctr. & Vel. & Qry. & $+$FPs &       &  \\
                \midrule
                --    & \multicolumn{4}{c|}{Baseline w/o DN} & 0.493 & 1.208 \\
                \midrule
                1     &       &       &       & \checkmark & 0.495 & 1.214 \\
                2     &       & \checkmark & \checkmark & \checkmark & 0.505 & 1.205 \\
                3     & \checkmark &       &       & \checkmark & 0.503 & 1.204 \\
                4     & \checkmark & \checkmark & \checkmark &       & 0.495 & 1.223 \\
                5     & \checkmark & \checkmark & \checkmark & \checkmark & \textbf{0.515} & 1.206 \\
                \midrule
                6     & \checkmark &       & \checkmark & \checkmark & 0.498 & 1.198 \\
                7     & \checkmark & \checkmark &       & \checkmark & 0.501 & \textbf{1.196} \\
                \bottomrule
            \end{tabular}
            \label{tab:abl_dn_targets}
        }
    \end{minipage}
    \addtocounter{table}{-1}
    \caption{\textbf{Ablation studies on denoising groups and targets.} 
    (a) Impact of the number of denoising groups and the grouping strategies. 
    Our proposed Hybrid grouping achieves the best performance. 
    (b) Effectiveness of different noise targets. 
    Combining geometric, motion, and instance-level noises yields the optimal AMOTA.
    Abbreviations: Geo. (Geometric), Inst. (Instance-level), Ctr. (Center), Vel. (Velocity), Qry. (Query feature), +FPs (Adding false positives), Gen. (General), Ded. (Dedicated), Hyb. (Hybrid).
    }
    \label{tab:combined_ablations}
\end{table}

We classify the noise types into three categories: the geometric noise such as 3D center noise, the motion noise including BEV velocity and query feature noise, and instance-level noise such as the random addition of previous-frame false positives. 
Table \ref{tab:abl_dn_targets} compares the impact of different combinations of these noise categories.
When only considering the instance-level noise (Exp. 1) or only considering geometric and motion noises (Exp. 4), the model achieves similar tracking performance as the baseline, suggesting that a better combination of denoising targets is needed.
Adding only geometric (Exp. 3) or motion (Exp. 2) noises on top of instance-level noises leads to a notably higher AMOTA.
By adding all three categories together, the model achieves highest AMOTA of \SI{0.515}{}.
This observation confirms our choice of introducing the three different noise categories.
In addition, when decoupling both noise types within motion noise (Exp. 6 and 7), the performance drops.
This indicates that both the velocity and the query feature representation are closely related to each other by modeling changes in the object motion.
This highlights the importance of introducing noise on both targets simultaneously.

\subsubsection{Noise Scales Tuning}
Three types of noise are added to the denoising queries, including uniform noise for the 3D center as Eq.~\eqref{equ:centernoise}, Gaussian noise for the BEV velocity as Eq.~\eqref{eq:velonoise}, and Gaussian noise for the query feature as Eq.~\eqref{eq:appearancenoise}.
Each type of noise has a corresponding noise scale which controls the noise intensity.
For the uniform noise added to the center, the scale is controlled by $\lambda$; for the Gaussian noise added to the BEV velocity and the query feature, the scale is controlled by $\sigma_\text{velo}$ and $\sigma_\text{query}$, respectively.
We set $\boldsymbol{\Sigma}^\text{velo} = \sigma_\text{velo} \mathbf{I} \in \mathbb{R}^{2 \times 2}$ and $\boldsymbol{\Sigma}^\text{query} = \sigma_\text{query} \mathbf{I} \in \mathbb{R}^{d_k \times d_k}$, where $d_k$ is the feature dimension and $\mathbf{I}$ is the identity matrix.
The larger the noise scale, the more noisy the denoising queries will be.
Table~\ref{tab:noisescales} shows the ablation study on the above mentioned noise scales.
Empirically we set $\lambda = 1$, $\sigma_\text{velo} = 4$ and $\sigma_\text{query} = 0.1$ in our experiments due to their best performance in the ablation study.
\bmvc{
The same hyperparameters are directly applied to AV2 experiments and yield consistent performance gains (Tab.~\ref{tab:per_class_hota}).}

\begin{table}[t!]
    \centering
    \stepcounter{table}
    \scriptsize 
    \setlength{\tabcolsep}{0.3mm}

    \begin{minipage}{0.54\textwidth} 
        \centering
        \subtable[\textbf{Noise scales.}]{
            \begin{tabular}{ccc|ccccc}
                \toprule
                $\lambda$ & $\sigma_\text{velo}$ & $\sigma_\text{query}$ & AMOTA$\uparrow$ & AMOTP$\downarrow$ & Recall$\uparrow$ & MOTA$\uparrow$ & IDS$\downarrow$ \\
                \midrule
                \rowcolor[rgb]{ .816, .808, .808} 0.8   & \cellcolor[rgb]{ 1, 1, 1}4 & \cellcolor[rgb]{ 1, 1, 1}0.1 & \cellcolor[rgb]{ 1, 1, 1}0.502 & \cellcolor[rgb]{ 1, 1, 1}1.207 & \cellcolor[rgb]{ 1, 1, 1}0.616 & \cellcolor[rgb]{ 1, 1, 1}0.462 & \cellcolor[rgb]{ 1, 1, 1}595 \\
                \rowcolor[rgb]{ .816, .808, .808} 1.2   & \cellcolor[rgb]{ 1, 1, 1}4 & \cellcolor[rgb]{ 1, 1, 1}0.1 & \cellcolor[rgb]{ 1, 1, 1}0.505 & \cellcolor[rgb]{ 1, 1, 1}\textbf{1.202} & \cellcolor[rgb]{ 1, 1, 1}0.597 & \cellcolor[rgb]{ 1, 1, 1}0.455 & \cellcolor[rgb]{ 1, 1, 1}\textbf{525} \\
                \midrule
                1     & \cellcolor[rgb]{ .816, .808, .808}2 & 0.1   & 0.506 & 1.203 & 0.595 & 0.459 & 587 \\
                1     & \cellcolor[rgb]{ .816, .808, .808}6 & 0.1   & 0.495 & 1.210  & 0.602 & 0.445 & 583 \\
                \midrule
                1     & 4     & \cellcolor[rgb]{ .816, .808, .808}0.05 & 0.495 & 1.207 & 0.603 & 0.453 & 590 \\
                1     & 4     & \cellcolor[rgb]{ .816, .808, .808}0.2 & 0.499 & \textbf{1.202} & 0.586 & 0.45  & 603 \\
                \midrule
                \rowcolor[rgb]{.816, .808, .808} \textbf{1}     & \textbf{4}     & \textbf{0.1}   & \textbf{0.515} & 1.206 & \textbf{0.623} & \textbf{0.467} & 596 \\
                \bottomrule
            \end{tabular}
            \label{tab:noisescales}
        }
    \end{minipage}
    \hspace{-3mm} 
    \begin{minipage}{0.42\textwidth}
        \centering
        \subtable[\textbf{Denoising query composition.}]{
            \begin{tabular}{c|ccccc}
                \toprule
                DN Query & AMOTA$\uparrow$ & AMOTP$\downarrow$ & Recall$\uparrow$ & MOTA$\uparrow$ & IDS$\downarrow$ \\
                \midrule
                Track Query & \textbf{0.515} & \textbf{1.206} & 0.623 & \textbf{0.467} & \textbf{596} \\
                \midrule
                All-zeros & 0.500 & 1.210 & \textbf{0.627} & 0.454 & 627 \\
                \bottomrule
                \multicolumn{6}{c}{} \\ 
            \end{tabular}
            \label{tab:abl_dn_query}
        }
    \end{minipage}
    \addtocounter{table}{-1}
    \caption{\textbf{Ablation studies on denoising configurations.} (a) Impact of noise scales. $\lambda$, $\sigma_{\text{velo}}$, and $\sigma_{\text{query}}$ control the intensity of center shifting, velocity noise, and query feature noise, respectively. (b) Comparison of different denoising query compositions. Initializing denoising queries with historical track query features outperforms all-zero initialization. 
    }
    \label{tab:merged_horizontal_ablations}
\end{table}
\subsubsection{Denoising Query Composition}
As shown in Eq.~\eqref{eq:appearancenoise}, the denoising query features are generated by adding Gaussian noise to the corresponding track query features.
To further validate our choice of aligning temporal  denoising queries with track queries w.r.t. their composition, we replace the denoising query features with all-zero vectors to remove the appearance cues encoded in the track query features, and report the results in Table~\ref{tab:abl_dn_query}. 
This leads to a notable \SI{1.5}{\%pt} decrease in AMOTA compared to \name{}, as the denoising queries no longer carry instance-specific information encoded in track query features.
Even in this case, our method still outperforms the baseline ADA-Track with an AMOTA of 0.493 and static denoising, further confirming our design choice in adopting query denoising in a streaming manner for MOT.

\subsection{Insights into Tracking Gains from TQD}
\label{suppl:learning_curve_analysis}
\begin{figure*}[t!]
    \centering
    \makeatletter
    \begin{minipage}[t]{0.58\textwidth}
        \centering
        {\scriptsize
        \setlength{\tabcolsep}{1.2mm}
        \renewcommand{\arraystretch}{0.92}
        \newcommand{\impcell}{\cellcolor[rgb]{0.84,0.96,0.84}}
        \resizebox{\linewidth}{!}{
        \begin{tabular}{lrrrrrrrr}
        \toprule
        & Trailer & Bus & Motocy & Bicy & Truck & Car & Ped. & \textbf{Agg.} \\
        \midrule
        $\Delta$IDS$\downarrow$  & $+$4 & \impcell$-$1 & $+$4 & \impcell$-$1 & \impcell$-$5 & \impcell$-$66 & \impcell$-$11 & \impcell\textbf{$-$76} \\
        $\Delta$FRAG$\downarrow$ & \impcell$-$5 & $+$9 & \impcell$-$6 & $\pm$0 & \impcell$-$9 & \impcell$-$20 & $+$5 & \impcell\textbf{$-$26} \\
        $\Delta$FP$\downarrow$   & \impcell$-$30 & $+$127 & $+$11 & \impcell$-$24 & $+$106 & \impcell$-$384 & \impcell$-$491 & \impcell\textbf{$-$685} \\
        \bottomrule
        \end{tabular}}
        \centerline{\scriptsize PETR-based ADA-Track on nuScenes val; Green denotes improvement.}}
        \refstepcounter{table}
        \label{tab:tracking_failure_analysis}
        \@makecaption{Table \thetable}{\textbf{Tracking error analysis.}}
    \end{minipage}
    \hfill
    \begin{minipage}[t]{0.38\textwidth}
        \centering
        \includegraphics[width=0.95\linewidth]{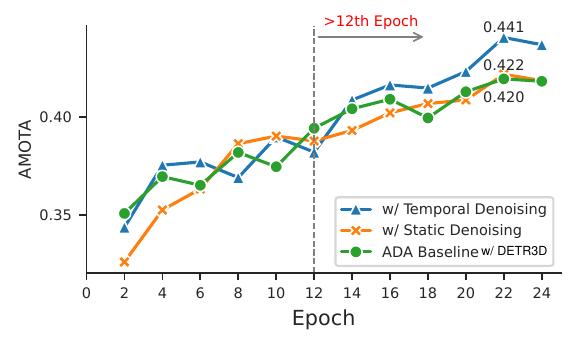}
        \caption{\textbf{Training curves.}}
        \label{fig:epoch}
    \end{minipage}
    \makeatother
\end{figure*}

\ready{We analyze how TQD improves tracking using both tracking error analysis (Table~\ref{tab:tracking_failure_analysis}) and training curves (Fig.~\ref{fig:epoch}).
The per-class breakdown is computed on the nuScenes val set with PETR-based ADA-Track, where $\Delta$ denotes the change after applying TQD.
Overall, TQD reduces IDS/FRAG/FP errors by 76/26/685, respectively.
Specifically, ID switches (IDS) are reduced from 672 to 596, and fragmentations (FRAG) are reduced from 995 to 969, indicating stronger association consistency.
The largest IDS reduction occurs for Car ($-66$), the densest category with frequent inter-frame ambiguity.
TQD also reduces false positives (FP) by 685, suggesting that our instance-level negative DN sampling helps the tracker suppress its own accumulated errors.
The learning curves in Fig.~\ref{fig:epoch} further explain why this training strategy is effective.
After 12 epochs, the baseline and static DN start to plateau, while TQD continues to improve and achieves the best final AMOTA.
We hypothesize that, as training progresses, high-quality track queries make next-frame association less challenging for the baseline and static DN, leading to performance saturation.
In contrast, TQD keeps providing challenging DN samples, further improving association in later training.
These results provide concrete insights into the gains of TQD and validate our design choices.
}

\begin{figure*}[t!]
    \centering
    \subfigure[\textbf{Occlusion Scenario.}]{%
        \includegraphics[width=0.49\textwidth]{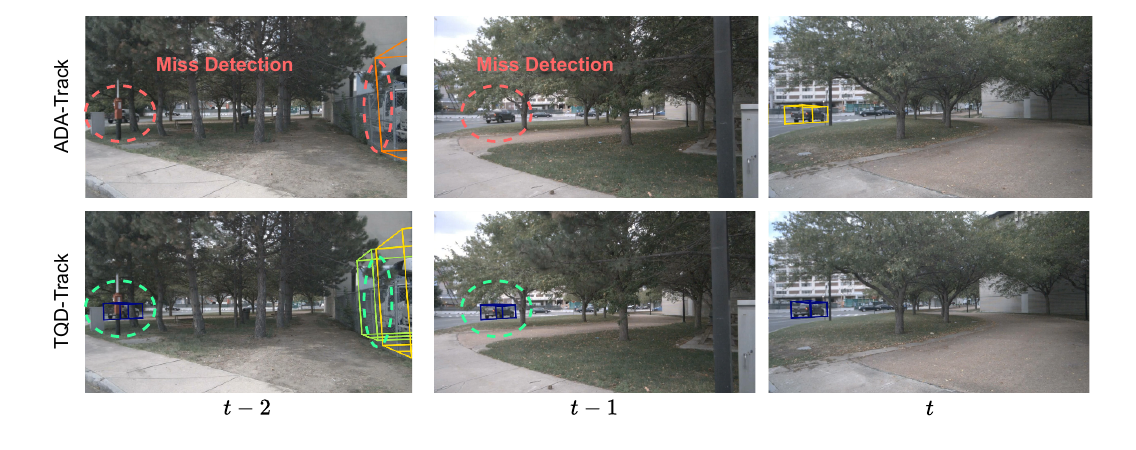}%
        \label{subfig:vis_fns}%
    }%
    \hspace{-3mm}
    \subfigure[\textbf{ID-Switch Scenario.}]{%
        \includegraphics[width=0.49\textwidth]{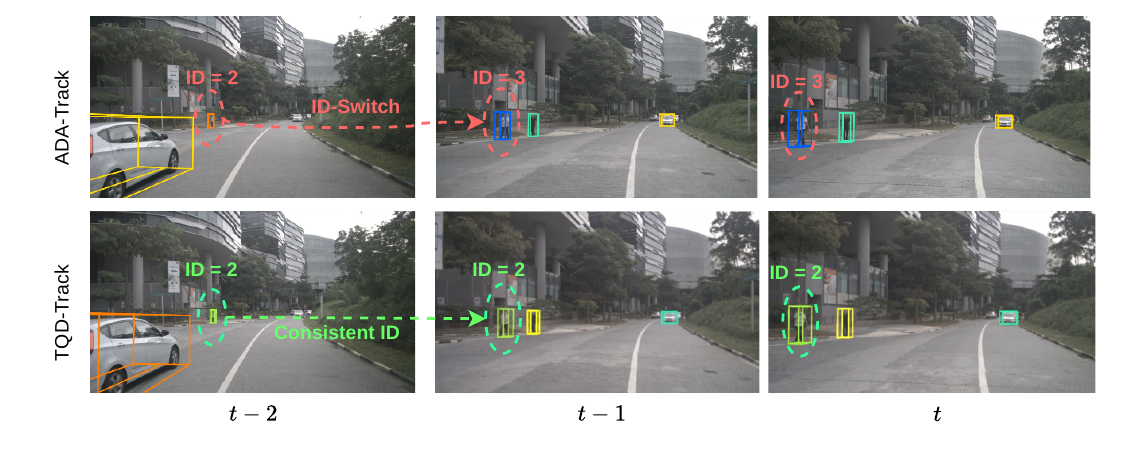}%
        \label{subfig:vis_ids}%
    }
    \caption{
    \textbf{Qualitative results on nuScenes.} TQD-Track handles (a) occlusions and (b) ID-switches more reliably than ADA-Track, demonstrating superior robustness.    }
    \label{fig:vis}
\end{figure*}

\subsection{Visualizations}
\bmvc{
Fig.~\ref{fig:vis} compares TQD-Track with ADA-Track~\cite{ding2024ada} in two challenging scenarios.
TQD-Track successfully handles missed detections caused by occlusion (Fig.~\ref{subfig:vis_fns}) and avoids identity switches (Fig.~\ref{subfig:vis_ids}) where the baseline fails. Notably, these gains are achieved with the same model complexity and inference speed.
}

\section{Conclusion}
\label{sec:conclusion}
We propose \name{}, a query denoising method tailored for 3D multi-object tracking (MOT).
Targeting the streaming nature of MOT, we introduce a temporal denoising strategy where denoising queries are generated from the previous frame and propagated to the current frame, serving as challenging training samples for association learning.
\bmvc{Comprehensive evaluations on both the nuScenes and Argoverse 2 benchmarks} demonstrate that temporal denoising consistently outperforms both vanilla training and standard static query denoising across different tracking paradigms,  highlighting its robustness and generalization capability.

\section*{Acknowledgement}
This work was supported by the German Federal Ministry for Economic Affairs and Energy (BMWE) based on a decision of the German Bundestag. 
Specifically, parts of this research were conducted within the joint project NXT GEN AI METHODS (nxtAIM), and parts within the joint project STADT:up (Förderkennzeichen 19A22006O).
The authors would like to thank both project consortia for the successful cooperation. 
The authors are solely responsible for the content of this publication.

\noindent Juergen Gall has been supported by the Deutsche
Forschungsgemeinschaft (DFG, German Research Foundation)
GA 1927/5-2 (FOR 2535 Anticipating Human
Behavior) and the ERC Consolidator Grant FORHUE
(101044724).

\bibliography{egbib}

@ARTICLE{ding2024adapp,
  author={Ding, Shuxiao and Schneider, Lukas and Cordts, Marius and Gall, Juergen},
  title={{ADA-Track++: End-to-End Multi-Camera 3D Multi-Object Tracking With Alternating Detection and Association}},
  journal={IEEE Transactions on Pattern Analysis and Machine Intelligence},
  year={2025},
  doi={10.1109/TPAMI.2025.3613269}
}

@inproceedings{carion2020end,
  title={End-to-end object detection with transformers},
  author={Carion, Nicolas and Massa, Francisco and Synnaeve, Gabriel and Usunier, Nicolas and Kirillov, Alexander and Zagoruyko, Sergey},
  booktitle={European conference on computer vision (ECCV)},
  pages={213--229},
  year={2020},
  organization={Springer}
}

@inproceedings{li2022dn,
  title={Dn-detr: Accelerate detr training by introducing query denoising},
  author={Li, Feng and Zhang, Hao and Liu, Shilong and Guo, Jian and Ni, Lionel M and Zhang, Lei},
  booktitle={IEEE/CVF Conference on Computer Vision and Pattern Recognition (CVPR)},
  pages={13619--13627},
  year={2022}
}

@inproceedings{zhangdino,
  title={DINO: DETR with Improved DeNoising Anchor Boxes for End-to-End Object Detection},
  author={Zhang, Hao and Li, Feng and Liu, Shilong and Zhang, Lei and Su, Hang and Zhu, Jun and Ni, Lionel and Shum, Heung-Yeung},
  booktitle={International Conference on Learning Representations (ICLR)},
  year={2023},
}

@inproceedings{liu2022petr,
  title={Petr: Position embedding transformation for multi-view 3d object detection},
  author={Liu, Yingfei and Wang, Tiancai and Zhang, Xiangyu and Sun, Jian},
  booktitle={European Conference on Computer Vision (ECCV)},
  pages={531--548},
  year={2022},
  organization={Springer}
}

@inproceedings{wang2022detr3d,
  title={Detr3d: 3d object detection from multi-view images via 3d-to-2d queries},
  author={Wang, Yue and Guizilini, Vitor Campagnolo and Zhang, Tianyuan and Wang, Yilun and Zhao, Hang and Solomon, Justin},
  booktitle={Conference on Robot Learning (CoRL)},
  pages={180--191},
  year={2022},
  organization={PMLR}
}

@inproceedings{caesar2020nuscenes,
  title={nuscenes: A multimodal dataset for autonomous driving},
  author={Caesar, Holger and Bankiti, Varun and Lang, Alex H and Vora, Sourabh and Liong, Venice Erin and Xu, Qiang and Krishnan, Anush and Pan, Yu and Baldan, Giancarlo and Beijbom, Oscar},
  booktitle={IEEE/CVF Conference on Computer Vision and Pattern Recognition (CVPR)},
  pages={11621--11631},
  year={2020}
}

@article{bernardin2008evaluating,
  title={Evaluating multiple object tracking performance: the clear mot metrics},
  author={Bernardin, Keni and Stiefelhagen, Rainer},
  journal={EURASIP Journal on Image and Video Processing},
  volume={2008},
  pages={1--10},
  year={2008},
  publisher={Springer}
}

@InProceedings{ding2024ada,
    author    = {Ding, Shuxiao and Schneider, Lukas and Cordts, Marius and Gall, Juergen},
    title     = {ADA-Track: End-to-End Multi-Camera 3D Multi-Object Tracking with Alternating Detection and Association},
    booktitle = {IEEE/CVF Conference on Computer Vision and Pattern Recognition (CVPR)},
    month     = {June},
    year      = {2024},
    pages     = {15184-15194}
}

@inproceedings{lee2019energy,
author = {Lee, Youngwan and Hwang, Joong-won and Lee, Sangrok and Bae, Yuseok and Park, Jongyoul},
title = {An Energy and GPU-Computation Efficient Backbone Network for Real-Time Object Detection},
booktitle = {IEEE/CVF Conference on Computer Vision and Pattern Recognition Workshops (CVPRW)},
month = {June},
year = {2019}
}

@inproceedings{Hussain2021EdgeaugmentedGT,
  title={Global self-attention as a replacement for graph convolution},
  author={Hussain, Md Shamim and Zaki, Mohammed J and Subramanian, Dharmashankar},
  booktitle={28th ACM SIGKDD Conference on Knowledge Discovery and Data Mining},
  pages={655--665},
  year={2022}
}

@inproceedings{Weng20193DMT,
  title={3d multi-object tracking: A baseline and new evaluation metrics},
  author={Weng, Xinshuo and Wang, Jianren and Held, David and Kitani, Kris},
  booktitle={IEEE/RSJ International Conference on Intelligent Robots and Systems (IROS)},
  pages={10359--10366},
  year={2020},
  organization={IEEE}
}

@inproceedings{meinhardt2022trackformer,
  title={Trackformer: Multi-object tracking with transformers},
  author={Meinhardt, Tim and Kirillov, Alexander and Leal-Taixe, Laura and Feichtenhofer, Christoph},
  booktitle={IEEE/CVF Conference on Computer Vision and Pattern Recognition (CVPR)},
  pages={8844--8854},
  year={2022}
}

@inproceedings{zeng2022motr,
  title={Motr: End-to-end multiple-object tracking with transformer},
  author={Zeng, Fangao and Dong, Bin and Zhang, Yuang and Wang, Tiancai and Zhang, Xiangyu and Wei, Yichen},
  booktitle={European Conference on Computer Vision (ECCV)},
  pages={659--675},
  year={2022},
  organization={Springer}
}

@article{xu2022transcenter,
  title={TransCenter: Transformers with dense representations for multiple-object tracking},
  author={Xu, Yihong and Ban, Yutong and Delorme, Guillaume and Gan, Chuang and Rus, Daniela and Alameda-Pineda, Xavier},
  journal={IEEE Transactions on Pattern Analysis and Machine Intelligence},
  volume={45},
  number={6},
  pages={7820--7835},
  year={2022},
  publisher={IEEE}
}

@inproceedings{zhao2022tracking,
  title={Tracking objects as pixel-wise distributions},
  author={Zhao, Zelin and Wu, Ze and Zhuang, Yueqing and Li, Boxun and Jia, Jiaya},
  booktitle={European Conference on Computer Vision (ECCV)},
  pages={76--94},
  year={2022},
  organization={Springer}
}

@inproceedings{Wang2023streampetr,
    author    = {Wang, Shihao and Liu, Yingfei and Wang, Tiancai and Li, Ying and Zhang, Xiangyu},
    title     = {Exploring Object-Centric Temporal Modeling for Efficient Multi-View 3D Object Detection},
    booktitle = {IEEE/CVF International Conference on Computer Vision (ICCV)},
    month     = {October},
    year      = {2023},
    pages     = {3621-3631}
}

@inproceedings{hu2023planning,
  title={Planning-oriented autonomous driving},
  author={Hu, Yihan and Yang, Jiazhi and Chen, Li and Li, Keyu and Sima, Chonghao and Zhu, Xizhou and Chai, Siqi and Du, Senyao and Lin, Tianwei and Wang, Wenhai and others},
  booktitle={IEEE/CVF Conference on Computer Vision and Pattern Recognition (CVPR)},
  pages={17853--17862},
  year={2023}
}

@inproceedings{jiang2023vad,
  title={Vad: Vectorized scene representation for efficient autonomous driving},
  author={Jiang, Bo and Chen, Shaoyu and Xu, Qing and Liao, Bencheng and Chen, Jiajie and Zhou, Helong and Zhang, Qian and Liu, Wenyu and Huang, Chang and Wang, Xinggang},
  booktitle={IEEE/CVF International Conference on Computer Vision (ICCV)},
  pages={8340--8350},
  year={2023}
}

@InProceedings{zhang2022mutr3d,
    author    = {Zhang, Tianyuan and Chen, Xuanyao and Wang, Yue and Wang, Yilun and Zhao, Hang},
    title     = {MUTR3D: A Multi-Camera Tracking Framework via 3D-to-2D Queries},
    booktitle = {IEEE/CVF Conference on Computer Vision and Pattern Recognition Workshops (CVPRW)},
    month     = {June},
    year      = {2022},
    pages     = {4537-4546}
}

@article{doll2023star,
  title={STAR-Track: Latent Motion Models for End-to-End 3D Object Tracking with Adaptive Spatio-Temporal Appearance Representations},
  author={Doll, Simon and Hanselmann, Niklas and Schneider, Lukas and Schulz, Richard and Enzweiler, Markus and Lensch, Hendrik PA},
  journal={IEEE Robotics and Automation Letters (RAL)},
  year={2023},
  publisher={IEEE}
}

@inproceedings{li2022time3d,
  title={Time3d: End-to-end joint monocular 3d object detection and tracking for autonomous driving},
  author={Li, Peixuan and Jin, Jieyu},
  booktitle={IEEE/CVF Conference on Computer Vision and Pattern Recognition (CVPR)},
  pages={3885--3894},
  year={2022}
}

@inproceedings{pang2023standing,
  title={Standing Between Past and Future: Spatio-Temporal Modeling for Multi-Camera 3D Multi-Object Tracking},
  author={Pang, Ziqi and Li, Jie and Tokmakov, Pavel and Chen, Dian and Zagoruyko, Sergey and Wang, Yu-Xiong},
  booktitle={IEEE/CVF Conference on Computer Vision and Pattern Recognition (CVPR)},
  pages={17928--17938},
  year={2023}
}

@inproceedings{li2023end,
  title={End-to-end 3D Tracking with Decoupled Queries},
  author={Li, Yanwei and Yu, Zhiding and Philion, Jonah and Anandkumar, Anima and Fidler, Sanja and Jia, Jiaya and Alvarez, Jose},
  booktitle={IEEE/CVF International Conference on Computer Vision (ICCV)},
  pages={18302--18311},
  year={2023}
}

@article{zhou2024ua,
  title={S2-track: A simple yet strong approach for end-to-end 3d multi-object tracking},
  author={Tang, Tao and Zhou, Lijun and Hao, Pengkun and He, Zihang and Ho, Kalok and Gu, Shuo and Hao, Zhihui and Sun, Haiyang and Zhan, Kun and Jia, Peng and others},
  journal={arXiv preprint arXiv:2406.02147},
  year={2024}
}

@inproceedings{fu2023denoising,
  title={Denoising-mot: Towards multiple object tracking with severe occlusions},
  author={Fu, Teng and Wang, Xiaocong and Yu, Haiyang and Niu, Ke and Li, Bin and Xue, Xiangyang},
  booktitle={Proceedings of the 31st ACM International Conference on Multimedia},
  pages={2734--2743},
  year={2023}
}

@article{zhu2020deformable,
  title={Deformable detr: Deformable transformers for end-to-end object detection},
  author={Zhu, Xizhou and Su, Weijie and Lu, Lewei and Li, Bin and Wang, Xiaogang and Dai, Jifeng},
  journal={arXiv preprint arXiv:2010.04159},
  year={2020}
}

@inproceedings{meng2021conditional,
  title={Conditional detr for fast training convergence},
  author={Meng, Depu and Chen, Xiaokang and Fan, Zejia and Zeng, Gang and Li, Houqiang and Yuan, Yuhui and Sun, Lei and Wang, Jingdong},
  booktitle={IEEE/CVF International Conference on Computer Vision (ICCV)},
  pages={3651--3660},
  year={2021}
}

@inproceedings{liudab,
  title={DAB-DETR: Dynamic Anchor Boxes are Better Queries for DETR},
  author={Liu, Shilong and Li, Feng and Zhang, Hao and Yang, Xiao and Qi, Xianbiao and Su, Hang and Zhu, Jun and Zhang, Lei},
  booktitle={International Conference on Learning Representations (ICLR)},
  year={2022}
}

@inproceedings{wang2024onetrack,
  title={OneTrack: Demystifying the Conflict Between Detection and Tracking in End-to-End 3D Trackers},
  author={Wang, Qitai and He, Jiawei and Chen, Yuntao and Zhang, Zhaoxiang},
  booktitle={European Conference on Computer Vision (ECCV)},
  pages={387--404},
  year={2024},
  organization={Springer}
}

@inproceedings{wang2024stream,
  title={Stream Query Denoising for Vectorized HD-Map Construction},
  author={Wang, Shuo and Jia, Fan and Mao, Weixin and Liu, Yingfei and Zhao, Yucheng and Chen, Zehui and Wang, Tiancai and Zhang, Chi and Zhang, Xiangyu and Zhao, Feng},
  booktitle={European Conference on Computer Vision (ECCV)},
  pages={203--220},
  year={2024},
  organization={Springer}
}

@article{wilson2023argoverse,
  title={Argoverse 2: Next generation datasets for self-driving perception and forecasting},
  author={Wilson, Benjamin and Qi, William and Agarwal, Tanmay and Lambert, John and Singh, Jagjeet and Khandelwal, Siddhesh and Pan, Bowen and Kumar, Ratnesh and Hartnett, Andrew and Pontes, Jhony Kaesemodel and others},
  journal={arXiv preprint arXiv:2301.00493},
  year={2023}
}

@article{luiten2021hota,
  title={Hota: A higher order metric for evaluating multi-object tracking},
  author={Luiten, Jonathon and Osep, Aljosa and Dendorfer, Patrick and Torr, Philip and Geiger, Andreas and Leal-Taix{\'e}, Laura and Leibe, Bastian},
  journal={International journal of computer vision},
  volume={129},
  number={2},
  pages={548--578},
  year={2021},
  publisher={Springer}
}

@article{chenle3de2e,
  title={Le3DE2E Solution for AV2 2024 Unified Detection, Tracking, and Forecasting Challenge},
  author={Chen, Feng and Lertniphonphan, Kanokphan and Meng, Yaqing and Ding, Ling and Xie, Jun and Huang, Kaer and Wang, Zhepeng}
}

@inproceedings{lin2026syncl,
  title={SynCL: A Synergistic Training Strategy with Instance-Aware Contrastive Learning for End-to-End Multi-Camera 3D Tracking},
  author={Lin, Shubo and Kou, Yutong and Wu, Zirui and Wang, Shaoru and Li, Bing and Hu, Weiming and Gao, Jin},
  booktitle={Advances in Neural Information Processing Systems (NeurIPS)},
  year={2025}
}

@article{song2026hypergraph,
  title={Hypergraph-state collaborative reasoning for multi-object tracking},
  author={Song, Zikai and Yu, Junqing and Chen, Yi-Ping Phoebe and Yang, Wei and Wang, Xinchao},
  journal={arXiv preprint arXiv:2604.12665},
  year={2026}
}

@inproceedings{song2025temporal,
  title={Temporal coherent object flow for multi-object tracking},
  author={Song, Zikai and Luo, Run and Ma, Lintao and Tang, Ying and Chen, Yi-Ping Phoebe and Yu, Junqing and Yang, Wei},
  booktitle={Proceedings of the AAAI Conference on Artificial Intelligence},
  volume={39},
  number={7},
  pages={6978--6986},
  year={2025}
}

@article{yang2026groupensemble,
  title={GroupEnsemble: Efficient Uncertainty Estimation for DETR-based Object Detection},
  author={Yang, Yutong and Popovi{\'c}, Katarina and Wiederer, Julian and Braun, Markus and Belagiannis, Vasileios and Yang, Bin},
  journal={arXiv preprint arXiv:2603.01847},
  year={2026}
}

@inproceedings{li2026spacedrive,
  title={Spacedrive: Infusing spatial awareness into vlm-based autonomous driving},
  author={Li, Peizheng and Zhang, Zhenghao and Holtz, David and Yu, Hang and Yang, Yutong and Lai, Yuzhi and Song, Rui and Geiger, Andreas and Zell, Andreas},
  booktitle={Proceedings of the IEEE/CVF Conference on Computer Vision and Pattern Recognition},
  pages={40096--40107},
  year={2026}
}

@inproceedings{li2025ago,
  title={Ago: Adaptive grounding for open world 3d occupancy prediction},
  author={Li, Peizheng and Ding, Shuxiao and Zhou, You and Zhang, Qingwen and Inak, Onat and Triess, Larissa and Hanselmann, Niklas and Cordts, Marius and Zell, Andreas},
  booktitle={2025 IEEE/CVF International Conference on Computer Vision (ICCV)},
  pages={8645--8655},
  year={2025},
  organization={IEEE}
}

@inproceedings{zhang2024seflow,
  title={Seflow: A self-supervised scene flow method in autonomous driving},
  author={Zhang, Qingwen and Yang, Yi and Li, Peizheng and Andersson, Olov and Jensfelt, Patric},
  booktitle={European Conference on Computer Vision},
  pages={353--369},
  year={2024},
  organization={Springer}
}

@inproceedings{du2026x,
  title={X-band Radar Non-Line-of-Sight Imaging},
  author={Du, Dongyu and Zhao, Mingkun and Yang, Yutong and Scheuble, Dominik and Huang, Xiaolong and Shao, Zijian and Bijelic, Mario and Sengupta, Kaushik and Heide, Felix},
  booktitle={Proceedings of the IEEE/CVF Conference on Computer Vision and Pattern Recognition},
  pages={5647--5658},
  year={2026}
}

\makesupplementheader

\noindent This Technical Appendix, a supplement to “\textbf{TQD-Track: Temporal Query Denoising for 3D Multi-Object Tracking}”, consists of:
\begin{itemize}
    \item Appendix~\ref{suppl:nuscenes_per_class}: Per-class Analysis on nuScenes
  \item Appendix~\ref{suppl:av2_details}: Implementation Details for Argoverse2
  \item Appendix~\ref{suppl:additional_ablation_studies}: Additional Ablation Studies on nuScenes
  \item Appendix~\ref{suppl:model_details}: Model Details and Association Mask
  \item Appendix~\ref{suppl:comp_complex}: Computation and Memory Complexity
  \item Appendix~\ref{suppl:tbd_results}: Tracking-by-Detection Paradigm Results
\end{itemize}

\setcounter{secnumdepth}{2}

\section{Per-class Analysis on nuScenes}
\label{suppl:nuscenes_per_class}
\begin{table}[H]
\centering
\label{tab:nuscenes_per_class}
\resizebox{\textwidth}{!}{
\begin{tabular}{l|ccccccc|c}
\toprule
Method & Trailer $\uparrow$ & Motor. $\uparrow$ & Bicycle $\uparrow$ & Truck $\uparrow$ & Bus $\uparrow$ & Car $\uparrow$ & Ped. $\uparrow$ & \textbf{Avg. AMOTA} $\uparrow$ \\
\midrule
ADA-Track \cite{ding2024ada} & 0.152 & 0.509 & 0.483 & 0.448 & 0.613 & 0.675 & \textbf{0.572} & 0.493 \\
+Static DN & 0.145 & 0.495 & 0.483 & 0.461 & \textbf{0.650} & 0.675 & 0.563 & 0.496 \\
\rowcolor[HTML]{EFEFEF} 
\textbf{+Temp DN (Ours)} & \textbf{0.243} & \textbf{0.532} & \textbf{0.505} & \textbf{0.463} & 0.618 & \textbf{0.678} & 0.569 & \textbf{0.515} \\
\bottomrule
\end{tabular}
}
\caption{Comparison of per-class AMOTA across different denoising strategies on nuScenes. }
\end{table}
\bmvc{To further validate the effectiveness of TQD-Track, we provide a per-class AMOTA breakdown on the nuScenes validation set (Tab.~\ref{tab:nuscenes_per_class}), evaluated using the PETR-based ADA-Track configuration.
Several key observations align with our findings on Argoverse 2:

\noindent \textbf{Robustness for Dynamic Objects.}
Categories with high mobility or highly variable kinematic patterns, such as \textit{Motorcycle} (+2.3\%pt) and \textit{Bicycle} (+2.2\%pt), consistently benefit from our temporal denoising. 
In contrast, static denoising often yields marginal or even negative gains on these categories, as it primarily augments single-frame box refinement (\ie, precise object localization), which misaligns with the inter-frame association task of MOT.

\noindent \textbf{Significant Gains in Challenging Categories.} TQD-Track achieves a notable +9.1\%pt gain in the challenging \textit{Trailer} category (improving AMOTA from 0.152 to 0.243), whereas static denoising results in a performance drop (-0.7\%). 
This contrast suggests that static denoising fails to generate effective training samples for MOT. TQD effectively bridges this gap by aligning its denoising process with the auto-regressive nature of MOT. 
By employing various temporally targeted noise types, TQD simulates diverse motion patterns and challenging association cases, thereby promoting robust tracking.

\noindent \textbf{Saturation in Common Categories.}
For well-behaved or frequently occurring categories like \textit{Car}, all three methods perform similarly (within a 0.3\%pt margin). 
This suggests that while standard trackers already handle common cases effectively, TQD demonstrates strong potential in resolving critical long-tail and high-uncertainty scenarios.}

\section{Implementation Details for Argoverse2}
\label{suppl:av2_details}

\subsection{Category Selection and Merging}
\bmvc{
Argoverse 2 (AV2) dataset~\cite{wilson2023argoverse} originally provides 26 fine-grained classes.
As multi-object tracking task primarily focuses on dynamic entities~\cite{zhang2022mutr3d,ding2024ada}, we first exclude 8 static or irrelevant categories from the 26 classes.
Given that several remaining dynamic categories exhibit similar semantic attributes and kinematic behaviors, \eg, various bus subtypes, we map the remaining 18 classes into 9 distinct classes. 
These 9 categories are used for both training and evaluation throughout our experiments.
Table~\ref{tab:av2_mapping} details the excluded categories and the mapping from original classes to their respective meta-classes.}

\begin{table}[htbp]
\centering
\label{tab:av2_mapping}
\resizebox{\linewidth}{!}{%
\begin{tabular}{ll}
\toprule
\textbf{Meta-class} & \textbf{Original Argoverse 2 Categories} \\ 
\midrule
\text{Regular Vehicle} & \texttt{REGULAR\_VEHICLE} \\
\text{Pedestrian} & \texttt{PEDESTRIAN} \\
\text{Bicycle} & \texttt{BICYCLE} \\
\text{Trailer} & \texttt{VEHICULAR\_TRAILER} \\
\text{Motorcycle} & \texttt{MOTORCYCLE} \\
\text{Personal Mobility} & \texttt{WHEELED\_DEVICE}, \texttt{WHEELCHAIR}, \texttt{STROLLER} \\
\text{Heavy Vehicle} & \texttt{LARGE\_VEHICLE}, \texttt{BOX\_TRUCK}, \texttt{TRUCK}, \texttt{TRUCK\_CAB} \\
\text{Bus} & \texttt{BUS}, \texttt{SCHOOL\_BUS}, \texttt{ARTICULATED\_BUS} \\
\text{Rider} & \texttt{BICYCLIST}, \texttt{MOTORCYCLIST}, \texttt{WHEELED\_RIDER} \\
\midrule
\textbf{Removal Reason} & \textbf{Excluded Categories} \\
\midrule
Static Signs & \texttt{STOP\_SIGN}, \texttt{SIGN}, \texttt{MOBILE\_PEDESTRIAN\_CROSSING\_SIGN} \\
Road Obstacles & \texttt{BOLLARD}, \texttt{CONSTRUCTION\_CONE}, \texttt{CONSTRUCTION\_BARREL} \\
Static Trailer & \texttt{MESSAGE\_BOARD\_TRAILER} \\
Animal & \texttt{DOG} \\
\bottomrule
\end{tabular}%
}
\caption{\textbf{Category mapping for Argoverse 2.} We consolidate 18 dynamic categories into 9 meta-classes and exclude 8 categories that are static or irrelevant.}
\end{table}

\subsection{DETR3D Detector Pretraining}
\label{suppl:detr3d_av2}

\bmvc{Following the experimental setup described in Sec. 4.1.3, all trackers are initialized with weights from a pre-trained detector. 
Specifically, for AV2 dataset, we utilize DETR3D~\cite{wang2022detr3d} equipped with a VoVNetV2~\cite{lee2019energy} backbone as the primary detector. 
The model is trained on the aforementioned 9 categories for 24 epochs.
Notably, both the training and evaluation stages are conducted within an ego-distance threshold of 50m, consistent with the official Argoverse 2 protocol.
Table~\ref{tab:detr3d_av2_results} reports the per-category detection performance in terms of Average Precision (AP).}
\begin{table}[htbp]
\centering
\label{tab:detr3d_av2_results}
\resizebox{\linewidth}{!}{%
\begin{tabular}{lcccccccccc}
\toprule
\textbf{Category} & Bicy. & Bus & Reg. Veh. & Heav. Veh. & Pers. Mob. & Moto. & Ped. & Rider & Trai. & \textbf{mAP} \\ \midrule
\textbf{AP ($\uparrow$)} & 31.0 & 43.7 & 62.0 & 33.3 & 19.6 & 42.9 & 35.5 & 29.6 & 20.5 & 35.3 \\ \bottomrule
\end{tabular}%
}
\caption{\textbf{Detection performance of pre-trained DETR3D on the AV2 validation set.} We report the Average Precision (AP \%) for each category and the mean AP (mAP).}
\end{table}

\begin{table}[t!]
\small
    \centering
    {%
    \begin{tabular}{cc|ccccc}
      \toprule
      $\alpha^\text{FP}$ & $\alpha^\text{drop}$ & AMOTA$\uparrow$ & AMOTP$\downarrow$ & Recall$\uparrow$ & MOTA$\uparrow$ & IDS$\downarrow$ \\
      \midrule
      \rowcolor[rgb]{ .816,  .808,  .808} 0     & \cellcolor[rgb]{ 1,  1,  1}0 & \cellcolor[rgb]{ 1,  1,  1}0.495 & \cellcolor[rgb]{ 1,  1,  1}1.223 & \cellcolor[rgb]{ 1,  1,  1}0.602 & \cellcolor[rgb]{ 1,  1,  1}0.452 & \cellcolor[rgb]{ 1,  1,  1}610 \\
      \rowcolor[rgb]{ .816,  .808,  .808} 0.05  & \cellcolor[rgb]{ 1,  1,  1}0 & \cellcolor[rgb]{ 1,  1,  1}0.505 & \cellcolor[rgb]{ 1,  1,  1}1.206 & \cellcolor[rgb]{ 1,  1,  1}0.609 & \cellcolor[rgb]{ 1,  1,  1}0.457 & \cellcolor[rgb]{ 1,  1,  1}\textbf{585} \\
      \rowcolor[rgb]{ .816,  .808,  .808} \textbf{0.1} & \cellcolor[rgb]{ 1,  1,  1}\textbf{0} & \cellcolor[rgb]{ 1,  1,  1}\textbf{0.515} & \cellcolor[rgb]{ 1,  1,  1}1.206 & \cellcolor[rgb]{ 1,  1,  1}0.623 & \cellcolor[rgb]{ 1,  1,  1}\textbf{0.467} & \cellcolor[rgb]{ 1,  1,  1}596 \\
      \rowcolor[rgb]{ .816,  .808,  .808} 0.15  & \cellcolor[rgb]{ 1,  1,  1}0 & \cellcolor[rgb]{ 1,  1,  1}0.503 & \cellcolor[rgb]{ 1,  1,  1}1.207 & \cellcolor[rgb]{ 1,  1,  1}0.604 & \cellcolor[rgb]{ 1,  1,  1}0.461 & \cellcolor[rgb]{ 1,  1,  1}578 \\
      \midrule
      0.1   & \cellcolor[rgb]{ .816,  .808,  .808}0.05 & 0.501 & 1.205 & 0.604 & 0.463 & 628 \\
      0.1   & \cellcolor[rgb]{ .816,  .808,  .808}0.1 & 0.507 & \textbf{1.199} & \textbf{0.624} & 0.458 & 654 \\
      \bottomrule
      \end{tabular}%
      }
    \vspace{5pt}
    \caption{Ablation study on the ratio of adding false positives $\alpha^\text{FP}$ and the ratio of randomly drop denoising samples $\alpha^\text{drop}$.}
    \label{tab:addfp}%
  \end{table}%

\section{Additional Ablation Studies on nuScenes}
\label{suppl:additional_ablation_studies}

\subsection{Instance-level Noise}
As described in Eq. 5, we add the Top-$k$ false positives from the detection queries during training to the denoising query set as negative denoising samples. These negative DN samples are supervised as background at the next frame after query propagation, teaching the auto-regressive tracker to correct its own previous prediction errors. 
In our method, $k$ is dynamically determined by a hyperparameter $\alpha^{FP}$, \ie, $k = \alpha^\text{FP} N_\text{GT}^{t-1}$, where $N_\text{GT}^{t-1}$ denotes the number of ground truth objects at frame $t-1$.
As a complementary study, we also investigate randomly dropping denoising queries using a drop ratio $\alpha^\text{Drop}$ which represents the probability that each positive denoising query is dropped.
Note that the random addition of false positives and the random dropping of queries are independent within each denoising group.
Table~\ref{tab:addfp} shows the evaluation of $\alpha^\text{FP}$ and $\alpha^\text{drop}$.
When the drop ratio $\alpha^\text{drop}$ is set to 0, increasing the false positive ratio $\alpha^\text{FP}$ improves the tracker's performance up to an optimal value of $\alpha^\text{FP} = 0.1$.
When further increasing $\alpha^\text{FP}$ to 0.15, the tracking performance decreases compared to $\alpha^\text{FP} = 0.1$, indicating that adding too many false positives could lead to data imbalance.
We then set false positive ratio $\alpha^\text{FP}$ to 0.1 and evaluate $\alpha^\text{drop}$.
As shown in the bottom part of Table~\ref{tab:addfp}, increasing the drop ratio $\alpha^\text{drop}$ results in a decrease in the tracker's performance compared to $\alpha^\text{drop} = 0$. 
This is because randomly dropping denoising queries reduces the number of effective denoising training samples available for the tracker to learn from.
Based on this ablation study, we set $\alpha^\text{FP} = 0.1$ and disable random dropping in our experiments.

\subsection{Number of Detection Queries}
The number of detection queries $N_\text{D}$ also affects the tracker's performance.
We conduct an ablation study on the number of detection queries on our \name{} and the results are shown in Table \ref{tab:numdet}.
The tracker achieves the best performance in terms of AMOTA when $N_\text{D} = 500$, and the Recall metric is consistently improved as the number of detection queries increases.
It's worth noting that our \name{} with 300 detection queries (49.9\% AMOTA) already outperforms the ADA-Track baseline with 500 detection queries (49.3\% AMOTA), which indicates the superiority of our proposed temporal query denoising method.  
  
  \begin{table}[t!]
  \small
    \centering
    \begin{tabular}{c|ccccc}
      \toprule
      $N_\text{D}$ & AMOTA$\uparrow$ & AMOTP$\downarrow$ & Recall$\uparrow$ & MOTA$\uparrow$ & IDS$\downarrow$ \\
      \midrule
      300   & 0.499 & 1.215 & 0.602 & 0.457 & 725 \\
      400   & 0.496 & \textbf{1.197} & 0.612 & 0.451 & 648 \\
      \textbf{500}   & \textbf{0.515} & 1.206 & \textbf{0.623} & \textbf{0.467} & \textbf{596} \\
      \bottomrule
      \end{tabular}%
    \vspace{5pt}
    \caption{Ablation study on the number of detection queries $N_\text{D}$ initialized at each frame.}
    \label{tab:numdet}%
  \end{table}%
\begin{figure}[t!]
  \centering
  \includegraphics[width=0.45\textwidth]{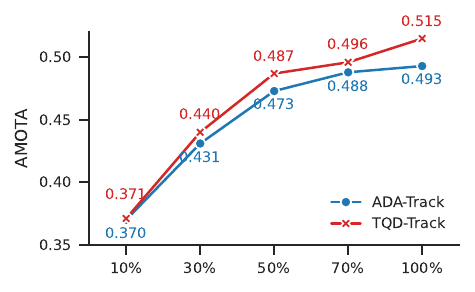}
  \caption{AMOTA vs. Portion of nuScenes training data by scaling the nuScenes training set.}
  \label{fig:portion}
\end{figure}

\subsection{Scalability to Training Data Volume}
We evaluate the scalability of \name{} to the training data volume by varying the amount of nuScenes training set.
In Figure~\ref{fig:portion}, we compare AMOTA of \name{} with ADA-Track with respects to the portions of the training set.
We start with only 10\% training set and both baseline and our \name{} achieve similar 37\% AMOTA.
By increasing the portions, \name{} consistently outperforms the baseline track across all the portion settings, and the performance gap between both methods tends to larger.
This could be attribute to the ability in augmenting diverse training samples of query denoising in \name{}, where more training data could lead to exponentially increased data diversity.
Notably, our TQD-Track achieves 49.6\% AMOTA when only using 70\% training set, which is better than the baseline tracker trained on the full training set (49.3\% AMOTA), showing its potential with limited training data.

\section{Model Details and Association Mask}
\label{suppl:model_details}
We integrated our temporal denoising method into multiple tracking paradigms, including tracking-by-attention (TBA) and alternating detection and association (ADA).
While TBA is a well-established and standard DETR-based tracker framework~\cite{meinhardt2022trackformer, zeng2022motr, zhang2022mutr3d}, ADA was introduced in a more recent work~\cite{ding2024ada}.
Therefore, we provide further details regarding ADA and its association mask for denoising in this section.
We also refer readers to ADA-Track~\cite{ding2024ada} for more comprehensive details.

\subsection{Details of ADA-Track}
\label{suppl:adatrack}
ADA-Track~\cite{ding2024ada} is a state-of-the-art multi-camera 3D MOT method which introduces the Alternating Detection and Association paradigm to overcome the weaknesses of both tracking-by-attention and tracking-by-detection paradigms.
At frame $t$, the multi-view images $I_{c}^{t}$ captured from surrounding $c$ cameras are first processed by a CNN image backbone to extract the feature maps $F_{c}^{t}$.
The input queries to the transformer decoder consist of two parts, namely the track queries $Q_\text{T}^{t}$ propagated from the previous frame and the detection queries $Q_\text{D}^{t}$ randomly initialized.
In ADA-Track, the track queries $Q_\text{T}^{t}$ are tasked to consistently detect the corresponding tracked objects, while the detection queries $Q_\text{D}^{t}$ are designed to detect all objects in the current frame $t$.
To encode the positional information of the queries, each query is assigned with a 3D reference point.
The reference points $C_\text{T}^{t}$ for track queries $Q_\text{T}^{t}$ are derived from the centers of the track boxes $\hat{B}_\text{T}^{t-1}$ from the previous frame $t-1$ after a motion update, while the reference points $C_\text{D}^{t}$ for detection queries $Q_\text{D}^{t}$ are initialized by uniformly sampling in the tracking space.
These two sets of queries as well as the edge features $E \in \mathbb{R}^{(N_\text{D} N_\text{T}) \times d_k}$ which are encoded from box differences between each track-detection query pair are also fed into the model with $L_{d}$ transformer decoder layers.
Each decoder layer applies a self-attention between queries before a cross-attention between queries and image features $F_{c}^{t}$.
Then, an Edge-Augmented Cross-Attention~\cite{Hussain2021EdgeaugmentedGT} module takes both query sets and the edges features as inputs to learn the data association between track and detection queries.

Formally (we omit the timestamp $t$ for simplicity here), the attention score of the Edge-Augmented Cross-Attention can be calculated as
\begin{equation}
\label{eq:eaca}
    A = \frac{(Q_\text{D} W_Q)(Q_\text{T} W_K)^T}{\sqrt{d_k}} + E W_{E1},
\end{equation}
where $\{ W_Q$, $W_K \} \in \mathbb{R}^{d_k \times d_k}$ and $W_{E1} \in \mathbb{R}^{d_k}$ are learnable weights.
The attention $A^{(l)} = \{ a^{(l)}_{ij} \}$ will be normalized among the second dimension $j$ using a softmax function, \ie
\begin{equation}
\label{eq:softmax1}
    \hat{a}_{ij} = \text{softmax}{( a_{ij} )}_j = \frac{\exp(a_{ij})}{ \sum_k^{N_\text{T}} \exp(a_{ik})}.
\end{equation}
The resulting normalized attention $\hat{A} = \{ \hat{a}_{ij} \}$ is used to update the detection query features by
\begin{equation}
\label{eq:update}
    Q_\text{D} \leftarrow Q_\text{D} + \hat{A}(Q_\text{T} W_V), \ Q_\text{D} \leftarrow Q_\text{D} + \text{FFN}_\text{D}(Q_\text{D})).
\end{equation}
In contrast, edge features are updated using $A$, \ie
\begin{equation}
\label{eq:update2}
    E \leftarrow E + A W_{E2}, \ E \leftarrow E + \text{FFN}_\text{E}(E)),
\end{equation}
where $W_V \in \mathbb{R}^{d_k \times d_k}$ and $W_{E2} \in \mathbb{R}^{d_k}$ denote learnable weights, $\text{FFN}_\text{D}$ and $\text{FFN}_\text{E}$ denote Feed-Forward-Networks (FFN) that update the detection query features and the edge features, respectively.
We omit the Layer Normalization in Eq.~\ref{eq:update} and Eq.~\ref{eq:update2} for simplicity.

In this way, the detection and association tasks are alternatively conducted across the $L_d$ decoder layers, forming the Alternating Detection and Association (ADA) paradigm.
At the end of the decoder, MLP-based task heads predict the bounding boxes $B_\text{T}^{t}$ and $B_\text{D}^{t}$ from the refined track queries $Q_\text{T}^{t}$ and detection queries $Q_\text{D}^{t}$, respectively.
Another MLP predict the association score $S$ between tracks and detections based on the final edge features $E^t$.
Finally, a track update module takes $B_\text{T}^{t}$ and $B_\text{D}^{t}$ along with their respective refined queries $Q_\text{T}^{t}$ and $Q_\text{D}^{t}$ as well as the association score $S$ as inputs, conducts an Hungarian Matching using $S$, and outputs the updated track queries $\hat{Q}_\text{T}^{t}$ and track boxes $\hat{B}_\text{T}^{t}$ based on the matching results.

\subsection{Association Mask in Temporal Denoising}
\bmvc{In our temporal denoising framework, denoising queries are treated as augmented track queries, forming a unified query set $Q_\text{U} = \bigl\{ Q_{\text{DN}, 1} \ , \ Q_{\text{DN}, 2} \ , \ ... \ , \ Q_{\text{DN}, N_\text{G}} \ , \ Q_\text{T} \bigr\}$. 
Since these denoising queries incorporate explicit ground-truth information, an association mask is essential to prevent information leakage from denoising queries to the track or detection queries within the explicit data association module of ADA-Track. 
This mask enables the model to effectively learn data association for denoising query samples without interfering with the feature representations of track or detection queries.}

Given unified denoising and track queries $Q_\text{U} \in \mathbb{R}^{(N_\text{DN,all}+N_\text{T}) \times d_k}$ as source and detection queries $Q_\text{D} \in \mathbb{R}^{N_\text{D} \times d_k}$ as target in cross-attention, the association mask $A_\text{asso} = \{ m^\text{asso}_{ij} \}$ with $i \in [0,N_\text{D})$ and $j \in [0, N_\text{DN,all}+N_\text{T})$ can be formulated by
\begin{equation}
\label{eq:asso_mask}
    m^\text{asso}_{ij} =
    \begin{cases}
        1 & \text{if} \  j < N_\text{DN,all}; \\
        0 & \text{otherwise}.
    \end{cases}
\end{equation}
Given cross-attention $a_{ij}$ between $i$-th query from $Q_\text{U}$ and $j$-th query from $Q_\text{D}$, we can use the association mask $m^\text{asso}_{ij}$ to calculate a masked softmax to normalize attention score solely among the track queries $Q_\text{T}$ from $Q_\text{U}$, \ie
\begin{equation}
\label{eq:softmax2}
    \hat{a}_{\text{mask},ij} = \sigma_\text{masked}(a_{ij}) = \frac{(1-m^\text{asso}_{ij}) \exp(a_{ij})}{\sum_k^{N_\text{DN,all}+N_\text{T}} (1-m^\text{asso}_{ik}) \exp(a_{ik})}.
\end{equation}
Using $\hat{A} = \{ \hat{a}_{\text{mask},ij} \}$, we update detection queries by only aggregating track query features by $Q_\text{D} \leftarrow Q_\text{D} + \hat{A}(Q_\text{T} W_V)$,
where $W_V$ denotes the learnable weights.

Different from query features, the features for association are established by each pair between $Q_\text{U}$ and $Q_\text{D}$, corresponding to the edge features in a graph that connects nodes (queries).
The association feature $E =\{e_{ij} \} \in \mathbb{R}^{{(N_\text{DN,all}+N_\text{T})N_\text{D} \times d_k}}$ between $i$-th source and $j$-th target nodes is generated by both node features $e_{ij} = f(q_i,q_j)$ or the attention $e_{ij}=f(a_{ij})$ or a combination of both where $f$ denotes the learned association function.
As the association features of each denoising-detection or track-detection pair are batched and stacked in a dimension, there is no information exchange between denoising-detection and track-detection pairs.
Therefore, no masking is needed for association features, which enables learning data association between denoising and detection queries.
To do so, the association feature is updated by the attentions $A$ before masked softmax normalization, \ie $E \leftarrow\ E + AW_{E}$.

The association mask allows denoising queries to effectively contribute to data association without GT leakage, thereby supporting MOT methods with explicitly learned data association modules.

\section{Computation and Memory Complexity}
\label{suppl:comp_complex}

\begin{table}[ht]
    \centering
    \label{tab:training_cost}
    \resizebox{\textwidth}{!}{
    \begin{tabular}{llccc}
        \toprule
        \textbf{Architecture} & \textbf{Method} & \textbf{AMOTA$\uparrow$} & \textbf{Training Time (h/epoch)} & \textbf{Training Memory (GB)} \\
        \midrule
        \multirow{3}{*}{MUTR3D~\cite{zhang2022mutr3d}} & Baseline & 0.443 & 1.21 & 5.5 \\
        & Static DN (5 DN groups) & 0.445 & 1.34 & 6.0 \\
        & Temp DN (5 DN groups) & \textbf{0.460} & 1.38 & 6.2 \\
        \midrule
        \multirow{3}{*}{ADA-Track~\cite{ding2024ada}} & Baseline & 0.493 & 1.20 & 15.2 \\
        & Static DN (5 DN groups) & 0.496 & 1.70 & 57.7$^\dagger$ \\
        & Temp DN (5 DN groups) & \textbf{0.515} & 1.74 & 62.6$^\dagger$ \\
        \bottomrule
        \multicolumn{5}{l}{\small $^\dagger$ The memory surge arises from processing the expanded query set through ADA-Track’s dense association module and wide FFNs (dim=2048).}
    \end{tabular}
    }
        \caption{Comparison of training costs per epoch and tracking performance across different architectures. All models are equipped with PETR~\cite{liu2022petr} as the detector. The training time is converted to hours per epoch (h/epoch) and training GPU memory to Gigabytes (GB). Note that the inference speed and model complexity remain completely unchanged.}
\end{table}

\bmvc{
As training augmentation strategies, both static and temporal query denoising introduce additional GPU memory consumption and training time due to the inclusion of extra training samples (\ie, denoising queries).
To evaluate this, we compare the training costs across different configurations in Table~\ref{tab:training_cost}.
Note that since our method is applied exclusively during training, the inference speed and model complexity remain unchanged.

First, we analyze the cost increase of static DN compared to the baselines.
The introduction of multiple denoising groups expands the total number of queries. 
For MUTR3D~\cite{zhang2022mutr3d}, the overhead is relatively modest (+0.13 h/epoch and +0.5 GB), as the query-level augmentation does not alter the model structure; the cost primarily stems from the enlarged self-attention affinity matrices within the decoder.
However, static DN incurs a significantly larger overhead for ADA-Track~\cite{ding2024ada} due to two architectural factors. 
First, its explicit graph-based association module requires computing pairwise edge features between detection queries and the expanded association candidates (\ie, the unified track-and-DN query set), as detailed in Appendix~\ref{suppl:model_details}.
Second, processing the expanded query set and the correspondingly enlarged edge features through ADA-Track's wide Feed-Forward Networks (FFNs with a hidden dimension of 2048 for both the association module and transformer decoder) substantially contributes to the training memory and time overhead.
Although reducing these FFN dimensions could alleviate the memory burden, we strictly adhere to the default configurations of ADA-Track for fair comparisons.

Next, we isolate the incremental cost introduced by temporal denoising by comparing our proposed Temporal DN (TQD) against static DN. 
Since the absolute training memory of ADA-Track is heavily influenced by architecture-specific components, particularly its wide association and decoder FFNs, comparing TQD against static DN provides a more informative comparison for assessing the extra cost of temporal modeling itself.
Under identical group settings, TQD introduces only a modest additional cost across both tracking architectures (\ie, +0.04 h/epoch and +0.2 GB for MUTR3D; +0.04 h/epoch and +4.9 GB for ADA-Track).
This overhead can be attributed to the introduction of instance-level noise (\ie, negative denoising samples). 
Unlike static DN, TQD explicitly samples high-scoring false positives (FPs) from the previous frame during training, and introduces them as negative DN queries to be supervised as background in the current frame. 
This strategy increases the total number of queries being processed, leading to slightly increased costs compared to static DN.
Importantly, however, it enforces the auto-regressive nature of the tracker to properly correct its previous errors by predicting previous-frame FPs as background during inference.

In summary, while denoising strategies inherently increase training costs to a certain degree depending on the underlying tracker architecture, TQD consistently improves tracking performance while maintaining a computational cost highly comparable to standard static DN.
This demonstrates that temporal denoising, equipped with diverse temporally targeted noise types, naturally aligns with the auto-regressive scheme of MOT for more effective training sample augmentation. 
}

\section{Tracking-by-Detection Paradigm Results}
\label{suppl:tbd_results}
Tracking-by-Detection (TBD) is a traditional tracking paradigm that first detect objects and then associate the detected objects with the tracks in the memory.
For completeness, we also report results for the Tracking-by-Detection (TBD) paradigm, following the TBD baseline implementation in ADA-Track~\cite{ding2024ada}.
Concretely, the TBD-baseline consists of a detector and a standalone data association decoder.
Track queries $Q_\text{T}^{t}$ and detection queries $Q_\text{D}^{t}$ with their reference points $C_\text{T}^{t}$, $C_\text{D}^{t}$ are fed into the detector.
The detector consists of transformer decoder layers with a self-attention between queries and a cross-attention between queries and image features, producing track and detection bounding boxes $B_\text{T}^{t}$ and $B_\text{D}^{t}$.
The association layer is same as in the ADA-Track, \ie Eq~\ref{eq:eaca}, Eq.~\ref{eq:softmax1}, Eq.~\ref{eq:update} and Eq.~\ref{eq:update2}.
However, they are not integrated into detector's decoder layer as in ADA-Track but are stacked together in the association decoder, corresponding to the modular design of tracking-by-detection.
The track update module remains the same as in ADA-Track.

  \begin{table}[t!]
  \small
    \centering
    \begin{tabular}{c|c|ccc}
    \toprule
    Paradigm & Denoising Method & AMOTA$\uparrow$ & AMOTP$\downarrow$ & Recall$\uparrow$ \\
    \midrule
    \multirow{3}[2]{*}{TBD} & w/o   & 0.483 & 1.249 & \textbf{0.603} \\
          & Static & 0.496 & 1.24  & 0.597 \\
          & Temporal (Ours) & \textbf{0.503} & \textbf{1.229} & 0.602 \\
    \bottomrule
    \end{tabular}%
      \label{tab:tbd}%
    \vspace{5pt}
    \caption{Comparisons of temporal denoising with no denoising and static denoising on the nuScenes validation set under the TBD paradigm.}
  \end{table}%

Following ADA-Track~\cite{ding2024ada}, we build the TBD baseline using a PETR~\cite{liu2022petr} detector followed by the learnable data association layers based on~\cite{Hussain2021EdgeaugmentedGT}.
VoVNetV2~\cite{lee2019energy} is used as the detector backbone.
Since the TBD baseline includes an explicit association module, both the self-attention mask and the association mask are applied when denoising is enabled.
In Table~\ref{tab:tbd}, we compare our proposed Temporal Query Denoising with Static Query Denoising and the vanilla training strategy under the TBD paradigm.
The results demonstrate that our temporal query denoising method consistently outperforms both the static denoising approach and the baseline without denoising, yielding a 2.0\%pt improvement in AMOTA over the TBD baseline w/o denoising and a 0.7\%pt gain over static denoising.
The similar trend can be observed for the ADA baseline in Table 1, which further validates the effectiveness of our design choice of integrating denoising queries into the data association module.

Across three tracking paradigms including Tracking-By-Attention (TBA), Tracking-By-Detection (TBD) and Alternating Detection and Association (ADA), our temporal query denoising method consistently and significantly outperforms both vanilla training and standard static query denoising, highlighting its generalization capability.
These gains can be attributed to the better alignment of temporal denoising with the streaming nature of the MOT task, enabling more efficient and effective training sample augmentation for the auto-regressive scheme in MOT.

\end{document}